\documentclass[preprint,12pt]{elsarticle}

\usepackage{lineno,hyperref}
\hypersetup{hidelinks}
\usepackage{bm}
\usepackage{amsmath}
\usepackage{amsfonts}       
\usepackage{amssymb}
\usepackage[linesnumbered,ruled]{algorithm2e} 
\usepackage{booktabs}       
\usepackage[section]{placeins}
\usepackage{subfigure}
\usepackage{multirow}
\usepackage{threeparttable}
\usepackage{color}
\graphicspath{{./fig/}}

\journal{Neurocomputing}
\date{}

\begin{document}

\begin{frontmatter}

\title{Semi-supervised Domain Adaptation on Graphs with Contrastive Learning and Minimax Entropy}

\author[mymainaddress]{Jiaren Xiao}
\author[mysecondaryaddress]{Quanyu Dai}
\author[mythirdaddress]{Xiao Shen}
\author[myfourthaddress,myfifthaddress]{Xiaochen Xie}
\author[mymainaddress]{Jing Dai}
\author[mymainaddress]{\\James Lam}
\author[mymainaddress]{Ka-Wai Kwok\corref{mycorrespondingauthor}}
\ead{kwokkw@hku.hk}
\cortext[mycorrespondingauthor]{Corresponding author}

\address[mymainaddress]{Department of Mechanical Engineering, The University of Hong Kong, Hong Kong, China}
\address[mysecondaryaddress]{Department of Computing, The Hong Kong Polytechnic University, Hong Kong, China}
\address[mythirdaddress]{School of Computer Science and Technology, Hainan University, Haikou, China}
\address[myfourthaddress]{Department of Automation, Harbin Institute of Technology, Shenzhen, China}
\address[myfifthaddress]{Institute for Automatic Control and Complex Systems, University of Duisburg-Essen, Duisburg, Germany}

\begin{abstract}
	Label scarcity in a graph is frequently encountered in real-world applications due to the high cost of data labeling. To this end, semi-supervised domain adaptation (SSDA) on graphs aims to leverage the knowledge of a labeled source graph to aid in node classification on a target graph with limited labels. SSDA tasks need to overcome the domain gap between the source and target graphs. However, to date, this challenging research problem has yet to be formally considered by the existing approaches designed for cross-graph node classification. This paper proposes a novel method called SemiGCL to tackle the graph \textbf{Semi}-supervised domain adaptation with \textbf{G}raph \textbf{C}ontrastive \textbf{L}earning and minimax entropy training. SemiGCL generates informative node representations by contrasting the representations learned from a graph's local and global views. Additionally, SemiGCL is adversarially optimized with the entropy loss of unlabeled target nodes to reduce domain divergence. Experimental results on benchmark datasets demonstrate that SemiGCL outperforms the state-of-the-art baselines on the SSDA tasks.
\end{abstract}

\end{frontmatter}

\section{Introduction}
\label{sec_intro}
Graphs represent the structured and relational data that are commonly found in the real world, such as social networks, citation networks, and protein-protein interaction (PPI) networks. Node classification is a crucial graph mining problem with practical applications in various fields like e-commerce and computational biology. Graph representation learning encodes graph information into low-dimensional node representation vectors, also known as node embeddings~\cite{hamilton_representation_2018}. These learned node representations can then be used with classical machine learning methods, such as a logistic regression classifier, to classify the unlabeled nodes. In recent years, graph neural networks (GNNs) have emerged as promising approaches for node classification tasks~\cite{wu_comprehensive_2021}. Furthermore, many recent GNN models apply graph contrastive learning to produce meaningful node representations in a self-supervised manner~\cite{liu_graph_2022}.

Existing studies primarily consider the node classification tasks on a partially labeled graph~\cite{wu_comprehensive_2021, cui_survey_2019}. In this single-graph scenario, a node classification model is trained with the labeled nodes in a graph, and evaluated on the unlabeled nodes within the same graph. However, in realistic situations, there is often a need to classify nodes in a newly collected graph (target graph) with a scarcity of node labels~\cite{shen_network_2021, wu_unsupervised_2020}. Due to the resource-intensive and time-consuming nature of data labeling~\cite{hu_strategies_2020}, it is desirable to transfer the knowledge of an available labeled graph (source graph) to assist in node classification on the target graph. For instance, on a newly-formed social network that lacks labels, it would be advantageous to transfer the abundant label information from a well-developed social network in order to classify users into interest groups. Similarly, the label information of a well-established citation database could be transferred to assign research topics to papers in a newly constructed citation network. In this cross-graph scenario, the source and target graphs can be treated as independent domains, namely the source domain and the target domain. Since the source and target graphs have distinct data distributions, there exists a domain divergence (or distribution shift) between these domains, hindering knowledge transfer across domains.

Domain adaptation aims to reduce the domain gap and improve a model's performance when deployed to the target domain~\cite{wilson_survey_2020, zhuang_comprehensive_2020}. The studies on domain adaptation mainly focus on computer vision (CV) and natural language processing (NLP), assuming that the data within each domain, such as images and text, are independent and identically distributed (i.i.d.). The majority of existing work considers unsupervised domain adaptation (UDA), where the target domain data are completely unlabeled~\cite{wilson_survey_2020}. In many practical applications, it is feasible to acquire a small amount of labeled data from the target domain. When utilized appropriately, these limited labeled target data can aid in model training in conjunction with the source data. This research problem is known as semi-supervised domain adaptation (SSDA)~\cite{li_ecacl_2021, saito_semi-supervised_2019}.

\begin{figure}[htbp]
	\centering
	\includegraphics[width=11.32cm]{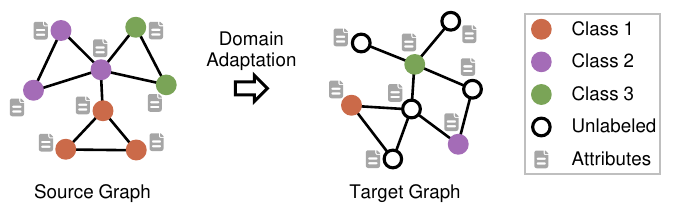}	
	\caption{Semi-supervised domain adaptation on graphs. The source and target graphs are two independent domains with distinct data distributions. The source graph is fully labeled, while the target graph has a limited number of labeled nodes per class.}
	\label{fig_task}
\end{figure}

This paper investigates the problem of cross-graph node classification under the SSDA setting, as depicted in Figure~\ref{fig_task}. The source graph is fully labeled, while the target graph has a limited number of labeled nodes per class. The majority of nodes (exceeding 99.5\% in most cases) are unlabeled in the target graph. The objective is to leverage information from both the source and target graphs to classify the unlabeled nodes in the target graph. Under the SSDA setting, it is crucial to effectively utilize the available labels in target graph to improve the model performance when predicting the unlabeled target nodes. One intuitive way is to optimize the model using the cross-entropy loss that takes both the labeled source and target nodes into account. However, without additional considerations, the node classification model would be biased towards the source graph with a much larger number of labeled nodes~\cite{li_ecacl_2021}. In addition, unlike images, nodes within a graph are connected by edges, thereby violating the i.i.d. assumption. It poses an additional challenge to the SSDA tasks.

Although domain adaptation in CV and NLP has been extensively studied, research on graph-structured data is still in its early stages. A few recent models (e.g., ACDNE~\cite{shen_adversarial_2020}, UDA-GCN~\cite{wu_unsupervised_2020}, ASN~\cite{zhang_adversarial_2021}, and MFRReg~\cite{you_graph_2023}) transfer knowledge from a labeled source graph to an unlabeled target graph with the help of domain adaptation techniques. Since the target graph is unlabeled, these models can be categorized into the UDA methods.

CDNE~\cite{shen_network_2021} and AdaGCN~\cite{dai_graph_2022} also explore the cross-graph node classification problem with the target graph partially labeled. However, they consider a target graph where a fraction of nodes (e.g., 5\% of the total nodes) are randomly selected to have accessible labels. In contrast, our work follows the commonly used SSDA setting~\cite{li_ecacl_2021, saito_semi-supervised_2019}. Under this setting, an equal number of labeled nodes (e.g., five nodes) are available for each class in the target graph. This setting would be more realistic and challenging. Given an originally unlabeled target graph, we can label a limited number of nodes per class to assist in the classification of each class node. To reduce the labeling effort, the number of labeled nodes in the target graph is much lower than 5\% of the total nodes, thereby increasing the difficulty of domain adaptation tasks. 

In addition, AdaGCN simply merges the labeled source and target nodes to calculate the cross-entropy loss. The domain adaptation technique adopted by AdaGCN is still an UDA one that reduces the Wasserstein distance between the source and target representations. CDNE, on the other hand, incorporates the target labels to calculate the class-conditional maximum mean discrepancy (MMD)~\cite{long_transfer_2013}. The reduction of MMD contributes to aligning the source and target distributions. However, the classical MMD metric has been empirically found inferior to some recent domain adaptation techniques~\cite{shen_wasserstein_2018}. Recent advances of domain adaptation in CV and NLP remain to be explored on the graph-structured data. For instance, the recently proposed SSDA approach, minimax entropy (MME)~\cite{saito_semi-supervised_2019}, adversarially optimizes an image classification model on the entropy loss of unlabeled target data, leading to improved model performance.

In this paper, we propose an end-to-end method called SemiGCL that addresses the graph \textbf{Semi}-supervised domain adaptation with \textbf{G}raph \textbf{C}ontrastive \textbf{L}earning and minimax entropy training. On the one hand, our method is empowered by graph contrastive learning to generate informative node representations for cross-graph node classification. Two graph neural network encoders jointly leverage graph structure information and node attributes to extract node representations from the original graph (local view) and the diffusion-augmented graph (global view), respectively. Graph contrastive learning is employed to maximize the mutual information between representations learned from the local and global views, which encourages the GNN encoders to capture a graph's rich local and global structural information simultaneously. 

On the other hand, as a design tailored to the SSDA setting, we introduce minimax entropy training to reduce the domain divergence between the source and target graphs. Our approach involves the utilization of a cosine similarity-based node classifier~\cite{chen_closer_2019}. The process of domain adaptation is modeled as a minimax game played between the GNN encoders and the node classifier. More precisely, the node classifier is trained to maximize the entropy of unlabeled nodes in the target graph. By doing so, the classifier’s weight vectors can be moved to the target graph, alleviating the model bias towards the source graph. In contrast, we update the GNN encoders to minimize the same entropy, which drives the node representations of unlabeled target nodes to get clustered around the classifier’s weight vectors, resulting in discriminative node representations. 

Our method is evaluated on eight benchmark transfer tasks using five real-world information networks. The \textit{main contributions} can be summarized as follows.
\begin{itemize}	
	\item The first GNN-based model is proposed and devised specifically to address the challenging semi-supervised domain adaptation problem on graphs.
	\item The model is designed to incorporate graph contrastive learning and minimax entropy training, effectively reducing domain divergence and generating discriminative node representations.
	\item Extensive validations are conducted on real-world information networks, demonstrating that our model outperforms the state-of-the-art approaches on the benchmark transfer tasks.
\end{itemize}

This paper is structured as follows. Section~\ref{sec_rel_work} provides a review of the related work. Section~\ref{sec_method} introduces the proposed method. Section~\ref{sec_exp} reports the experimental results. Finally, Section~\ref{sec_con} presents the conclusion.

\section{Related Work}
\label{sec_rel_work}
\subsection{Domain Adaptation}
Existing work mainly focuses on the UDA problem that assumes the data of target domain are purely unlabeled. The basic approach of UDA is to reduce the domain divergence through moment matching~\cite{long_transfer_2013, long_learning_2015, wang_deep_2020} or adversarial learning~\cite{shen_wasserstein_2018, ganin_domain-adversarial_2016, wang_structure-conditioned_2022, zhao_domain_2021}. Deep adaptation network (DAN)~\cite{long_learning_2015} estimates the distribution divergence with maximum mean discrepancy. The reduction of MMD helps match the source and target distributions. DANN~\cite{ganin_domain-adversarial_2016} introduces a domain classifier that is trained with the feature extractor in an adversarial manner. Specifically, the domain classifier is optimized to tell apart samples from the source and target domains. The feature extractor is trained to produce samples that can deceive the domain classifier, thus matching the feature distributions. Another adversarial adaptation method, WDGRL~\cite{shen_wasserstein_2018}, utilizes the Wasserstein distance as the adversarial loss.

Differing from the mainstream UDA research, a few recent studies investigate the SSDA problem where a small number of data are labeled in the target domain. Minimax entropy model~\cite{saito_semi-supervised_2019} consists of a feature encoder (e.g., AlexNet and VGG16) and a cosine similarity-based classifier~\cite{chen_closer_2019} devised for the few-shot classification. With the entropy of unlabeled target data calculated as adversarial loss, MME alleviates domain discrepancy by enforcing a minimax game between the feature extractor and the image classifier. ECACL~\cite{li_ecacl_2021} incorporates categorical alignment and consistency alignment to improve the performance of existing UDA methods under the SSDA setting.

Domain adaptation has been widely studied in areas like CV and NLP. However, most previous models are designed for vector-based data (e.g., images and text) that follow the i.i.d. assumption. Our work considers the domain adaptation problem for graph-structured data where nodes have complicated edge connections, thereby violating the i.i.d. assumption. Note that GRDA~\cite{xu_graph-relational_2022} studies the problem of domain adaptation across multiple domains with adjacency described by a domain graph. Each domain is a node in the domain graph. This problem is completely different from the one we considered, i.e., domain adaptation on graphs, where each domain is a graph. Therefore, GRDA is not applicable to our setting.

\subsection{Representation Learning on Graphs}
Graph representation learning generates low-dimensional representation vectors for graph nodes~\cite{hamilton_representation_2018, cui_survey_2019}. There are methods designed to preserve structural properties only~\cite{grover_node2vec_2016, perozzi_deepwalk_2014} or to jointly exploit graph structure and side information such as node attributes~\cite{xiao_adversarially_2021, zhang_anrl_2018, chen_simple_2020}. Among them, GNNs~\cite{wu_comprehensive_2021} have drawn a lot of research interest in recent years.

GraphSAGE~\cite{hamilton_inductive_2017} pioneers in generating node representations by aggregating information from the sampled neighboring nodes. Unlike GraphSAGE, which is evaluated in the supervised learning scenario, many studies focus on graph semi-supervised learning to overcome the label shortage in a partially labeled graph. The well-known model, GCN~\cite{kipf_semi-supervised_2017}, simplifies the spectral graph convolution as a first-order approximation. Through jointly encoding local graph structure and node attributes, GCN demonstrates impressive performance and inspires many follow-up studies. GAT~\cite{velickovic_graph_2018}, for instance, introduces an attention mechanism and assigns learnable weights to the neighbors. Note that GCN and GAT utilize all neighbors of a node without neighborhood sampling. 

Self-supervised learning~\cite{liu_graph_2022} is another way to tackle the lack of labels. It enables the models to produce meaningful node embeddings by conducting various pretext tasks without any node label information. Under this learning paradigm, graph contrastive learning (GCL) maximizes the mutual information (MI) between two (augmented) instances of the same object, such as one graph and a node. For example, DGI~\cite{velickovic_deep_2019} maximizes the MI between node representations and the global summaries of a graph. The resulting node representations are expected to encode the global structural information. MVGRL~\cite{hassani_contrastive_2020} contrasts the node and graph embeddings learned from two structural views of a graph.

The majority of graph representation learning models are not specially designed to address the domain discrepancy between graphs. Although we can sometimes adapt these models for cross-graph node classification, the domain divergence would still hinder their performance.

\subsection{Cross-graph Node Classification}
Recently, a few studies have been conducted to address the problem of cross-graph node classification, specifically focusing on two independent attributed graphs. The objective of this research is to assist in node classification on a target graph that lacks sufficient node labels by transferring knowledge from a source graph with abundant labels.

Similar to the UDA studies in CV and NLP, most existing methods are devised for a cross-graph scenario that assumes the target graph is unlabeled. To capture the attributed affinity and topological proximity, ACDNE~\cite{shen_adversarial_2020} constructs two feature extractors that encode node attributes and neighborhood attributes, respectively. UDA-GCN~\cite{wu_unsupervised_2020} utilizes dual graph convolutional networks to exploit the local and global consistency on a graph. Both ACDNE and UDA-GCN follow the adversarial training paradigm in DANN~\cite{ganin_domain-adversarial_2016} to match the embedding distributions of source and target graphs. ASN~\cite{zhang_adversarial_2021} also adopts the dual GCNs and improves the extraction of domain-shared features by separating the domain-private features. Xiao et al. recently proposed a GNN-based model (i.e., AdaGIn) that aligns the multimodal embedding distributions by conditional adversarial networks~\cite{xiao_domain_2022}. To capture the global graph information, AdaGIn maximizes mutual information between the node and graph-level representations. 

Very recently, a few studies design theory-grounded algorithms for domain adaptation on graphs. For example, MFRReg~\cite{you_graph_2023} regularizes the GNN spectral property (i.e., maximum frequency response) to improve the GNN transferability. Additionally, inspired by the Weisfeiler-Lehman graph isomorphism test, GRADE~\cite{wu_non-iid_2023} proposes a graph subtree discrepancy to measure the graph distribution shift. The reduction of graph subtree discrepancy contributes to aligning the graph data distributions. 

Unlike these UDA methods, CDNE~\cite{shen_network_2021} and AdaGCN~\cite{dai_graph_2022} consider a target graph in which a percentage of nodes are randomly selected to have accessible labels. CDNE generates node representations with stacked autoencoders. The source and target distributions are aligned by reducing the maximum mean discrepancy. AdaGCN extracts node representations with GCN and minimizes the Wasserstein distance~\cite{shen_wasserstein_2018} to reduce domain discrepancy. Our work focuses on the SSDA problem on graphs. To our knowledge, this problem has never been formally considered by the prior art. To address this problem, we devise a novel GNN-based model empowered by graph contrastive learning and minimax entropy training.
\section{Proposed Method}
\label{sec_method}
In this section, we first define the research problem. Then, the model architecture is presented. After that, we elaborate on node representation learning, node label prediction, and semi-supervised domain adaptation. Finally, we summarize the overall objective and outline the model training procedure.
\subsection{Problem Definition}
\label{sec_def_not}
An attributed graph, or an information network, is mathematically described as $ \mathcal{G}\left(\bm{V},\bm{A},\bm{X},\bm{Y}\right) $, including node set $ \bm{V}\in\mathbb{R}^{N} $, adjacency matrix $ \bm{A}\in\mathbb{R}^{N\times N} $, attribute matrix $ \bm{X}\in\mathbb{R}^{N\times L} $, and label matrix $ \bm{Y}\in\mathbb{R}^{N\times C} $. $ N $, $ L $, and $ C $ denote the number of nodes, the dimension of a node attribute vector, and the number of node classes, respectively. The $ i $-th rows of $ \bm{A} $, $ \bm{X} $, and $ \bm{Y} $ are associated with the $ i $-th node $ v\in\bm{V} $. Adjacency matrix $ \bm{A} $ and label matrix $ \bm{Y} $ contain binary values. Specifically, $ A_{ij}=1 $ indicates the $ i $-th node has an edge connection with the $ j $-th node; $ Y_{ic}=1 $ means the $ i $-th node belongs to the $ c $-th class. The degree of node $ v $ is the number of its edge connections, i.e., $ \sum\nolimits_{j} A_{ij} $. In this paper, we consider the undirected graph. The average degree is computed by $ \sum\nolimits_{i}\sum\nolimits_{j} A_{ij}/N $, revealing the graph density. Table~\ref{tab_notation} summarizes the main notations in this paper. 

\begin{table}[htbp]
	\caption{Main notations.}
	\label{tab_notation}
	\centering
	\centerline{
	\begin{tabular}{l|l}
		\toprule
		Notation                                              & Description                                                                            \\ \midrule
		$ \mathcal{G} $                                       & An attributed graph                                                                    \\
		$ \mathcal{G}^{s} $, $ \mathcal{G}^{t} $              & Source graph and target graph                                                          \\
		$ \bm{V} $, $ \bm{A} $                                            & Node set and adjacency matrix                                                          \\
		$ \bm{X} $, $ \bm{Y} $                              & Attribute matrix and label matrix      \\
		$ \bm{E} $, $ \hat{\bm{Y}} $                              & Representation matrix and label prediction matrix of $ \mathcal{G} $      \\
		$ \bm{\mathcal{X}} $                                  & Union attribute set \\
		$ \bm{x}_v $, $ \bm{y}_{v} $                              & Attribute vector and label vector of node $ v\in \bm{V} $       \\
		$ \bm{e}_{v} $, 		$ \hat{\bm{y}}_{v} $                          & Embedding vector and label prediction vector of node $ v\in \bm{V} $ \\
		$ n $                                                 & Number of labeled nodes per class in $ \mathcal{G}^{t} $                                                     \\ \midrule
		$ f_{A} $, $ f_{P} $, $ f_{c}  $                       & Local-view GNN encoder, global-view GNN encoder, and node classifier                      \\
		$ \bm{\theta}_{g}, \bm{\theta}_{c} $ & Sets of parameters in the GNN encoders and in the node classifier                              \\
		$ \alpha $                                                 & Teleport probability                                                               \\
		$ \sigma $                                            & Nonlinear activation function                                                          \\
		$ s $                                                 & Neighborhood sample size                                                               \\
		$ T $                                                 & Temperature parameter                                                               \\
		$ \eta_{0} $                                          & Initial learning rate                                                                  \\
		$ \bm{B} $                                            & A batch of nodes                                                                       \\
		$ \lambda_{1} $, $ \lambda_{2} $, $ \lambda_{3} $                                       & Contrastive learning coefficient, domain adaptation coefficient, and entropy coefficient                                                    \\ \bottomrule
	\end{tabular}}
\end{table}

Graph representation learning encodes each node $ v\in \bm{V} $ into a low-dimensional representation vector, that is, node embedding vector $ \bm{e}_{v} $. In representation matrix $ \bm{E}\in \mathbb{R}^{N\times l} $, $ \bm{e}_{v}^{\top} $ is stored as one row. The embedding dimension is denoted as $ l $. Using the learned node embeddings, a node classifier can be trained to perform node classification tasks. 

In this paper, we study the problem of semi-supervised domain adaptation on graphs. The source domain refers to a labeled graph $ \mathcal{G}^{s}\left(\bm{V}^{s},\bm{A}^{s},\bm{X}^{s},\bm{Y}^{s}\right) $, in which the label of each node is known. The target domain is a partially labeled graph $ \mathcal{G}^{t}\left(\bm{V}^{t},\bm{A}^{t},\bm{X}^{t},\bm{Y}^{t}\right) $, where only a few nodes have known labels. In line with the common SSDA setting~\cite{li_ecacl_2021, saito_semi-supervised_2019}, we randomly choose an equal number of nodes from each class to form the set of labeled nodes in the target graph. The objective of this study is to improve the node classification performance on the target graph by transferring knowledge from the source graph. To accomplish this, it is essential to produce node representations that are both transferrable and discriminative. This research problem is challenging because of the distinction between the source and target graphs. Since these two graphs have no edge connections or shared common nodes, they represent independent domains. The discrepancy between the domains arises from variations in graph scales and differences in the distributions of node attributes, edge connections, and node labels. 

Based on prior studies~\cite{shen_network_2021, wu_unsupervised_2020, dai_graph_2022}, it is necessary for the source and target graphs to have the same set of node classes. However, these two graphs possess different attribute sets, namely $ \bm{\mathcal{X}}^{s} $ and $ \bm{\mathcal{X}}^{t} $. A union attribute set is created and denoted as $ \bm{\mathcal{X}}=\bm{\mathcal{X}}^{s}\cup\bm{\mathcal{X}}^{t} $. The attribute matrices can then be expressed as $ \bm{X}^{s}\in\mathbb{R}^{N^{s}\times U} $ and $ \bm{X}^{t}\in\mathbb{R}^{N^{t}\times U} $, where $ U=\left| \bm{\mathcal{X}} \right| $ represents the total number of attributes across both graphs. This union attribute set enables parameter sharing, allowing the same model to be applied to both graphs. Many modern domain adaptation techniques rely on parameter sharing~\cite{saito_semi-supervised_2019, shen_wasserstein_2018, ganin_domain-adversarial_2016}. To quantify the extent of attribute overlap, we define a common attribute rate as $ R_{a}=\left| \bm{\mathcal{X}}^{s}\cap\bm{\mathcal{X}}^{t}\right|/\left| \bm{\mathcal{X}}\right| $, showing the percentage of node attributes present in both graphs.
\subsection{Overview of Model Architecture}
\begin{figure}[htbp]
	\centering
	\includegraphics[width=9.91cm]{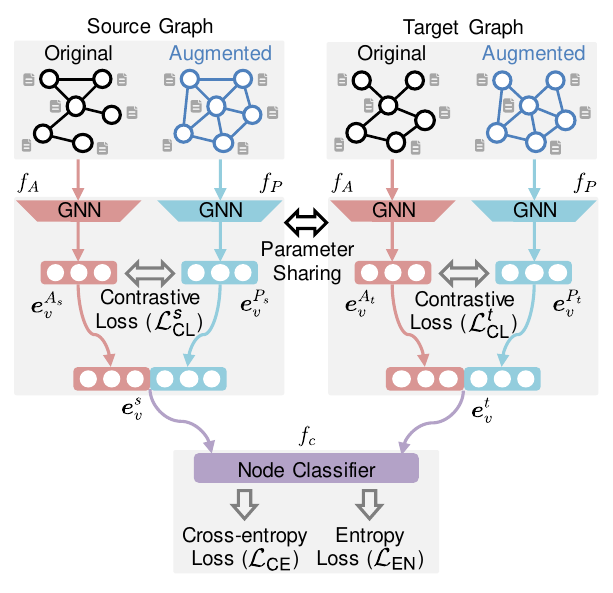}		
	\caption{Architecture of the proposed model. Two GNN encoders extract node representations from two structural views of a graph, i.e., the original graph (local view) and the diffusion-augmented graph (global view). Node representations from the local and global views are then contrasted and concatenated to obtain an informative node embedding vector. The cosine similarity-based node classifier computes the label prediction by taking the node embedding vector as input. The domain divergence between the source and target graphs is reduced by adversarially optimizing the model with the entropy loss. More details can be found in Section~\ref{sec_method}.}
	\label{fig_model_archi}
\end{figure}

As shown in Figure~\ref{fig_model_archi}, the proposed SemiGCL model is made up of two modules: the GNN encoders and the node classifier. In addition to using the original graph as a regular structural view, we use graph diffusion to obtain an augmented graph that serves as an additional structural view. Two GNN encoders are used to encode the original graph and the augmented graph, respectively. Representation vectors generated by the GNN encoders are then concatenated to obtain the node embedding vector. A cosine similarity-based classifier is employed as the node classifier, which takes the embedding vector as input and outputs the label prediction.

There are three losses involved in the model optimization. The contrastive loss is calculated for each graph in a self-supervised way. By minimizing the contrastive loss, the agreement is maximized between the representations learned from the two structural views. The cross-entropy loss is computed with the labeled nodes in the source and target graphs. The minimization of cross-entropy loss contributes to extracting discriminative node representations. The entropy loss is calculated on the unlabeled target nodes. To mitigate domain divergence, the GNN encoders and the node classifier are adversarially optimized with the entropy loss.

\subsection{Node Representation Learning}
We consider two structural views of the same graph: one regular view (local view) and one additional view (global view). The regular structural view is the original graph represented by binary adjacency matrix $ \bm{A} $. The adjacency matrix stores the edge connection, which is a kind of local structural information around a node. Hence, the regular view is also regarded as a local view. The additional structural view is an augmented graph described by diffusion matrix $ \bm{P} $. The augmented graph is generated using graph diffusion~\cite{klicpera_diffusion_2019}.
\begin{equation}
	\tilde{\bm{P}}=\left(\bm{D}+\bm{I}_{N}\right)^{-1/2}\left(\bm{A}+\bm{I}_{N}\right)\left(\bm{D}+\bm{I}_{N}\right)^{-1/2}
	\label{eq_trans}
\end{equation}
\begin{equation}
	\bm{P}=\alpha\left[\bm{I}_{N}-\left(1-\alpha\right)\tilde{\bm{P}}\right]^{-1}
	\label{eq_diff}
\end{equation}
where $ \bm{D} $ is the diagonal degree matrix of adjacency matrix $ \bm{A} $, $ \bm{I}_{N}\in\mathbb{R}^{N\times N} $ is an identity matrix, $ \tilde{\bm{P}} $ is the transition matrix, and $ \alpha\in\left(0,1\right) $ is the teleport probability. The edge weight in diffusion matrix represents the influence between two nodes that may not be in the one-hop neighborhood of the original graph. In other words, diffusion matrix can establish links to the distant nodes of original graph. The application of diffusion matrix makes nodes in multi-hop neighborhoods directly get involved in the node representation learning, capturing the global information. Therefore, the augmented structural view can be treated as a global view. Following GCN~\cite{kipf_semi-supervised_2017}, the calculation of diffusion matrix is regarded as a pre-processing step before loading the graph data to conduct model training.

\subsubsection{Representation Learning on Structural Views}
Node representations of the local and global views are calculated following the neighborhood aggregation strategy of modern GNNs~\cite{xu_how_2019}. Through aggregating neighborhood information based on the adjacency matrix and the diffusion matrix, the GNN encoders can naturally exploit both graph structure and node attributes to generate node representations. The node representation of local view is computed as follows.
\begin{equation}
	\bm{h}_{S}^{k}=\dfrac{1}{\left| \bm{S}_{v}\right|}\sum\nolimits_{u\in \bm{S}_{v}}\bm{h}_{u}^{k-1},
	\label{eq_neib}
\end{equation}
\begin{equation}
	\bm{h}_{v}^{k}=\sigma_{1}\left({\bm{W}_{A}^{k}[\bm{h}_{v}^{k-1};\bm{h}_{S}^{k}]}\right),
	\label{eq_concat}
\end{equation}
where the set $ \bm{S}_{v} $ represents a sample of nodes taken from the one-hop neighborhood of node $ v $, the vector $ \bm{h}_{S}^{k} $ is the neighborhood representation at step $ k\in\left\lbrace 1,\ldots,K \right\rbrace $, the vector $ \bm{h}_{v}^{k} $ is the representation of node $ v $ at step $ k $, $ \bm{W}_{A}^{k} $ is a weight matrix, and $ \sigma_{1} $ is the ReLU activation, that is, $ \sigma_{1}\left(x\right) ={\rm max}(0,x) $. Initially, node representation $ \bm{h}_{v}^{0} $ is set to be node attribute vector $ \bm{x}_{v} $. A simple form of skip connection~\cite{hamilton_inductive_2017} is implemented in Eq.~\ref{eq_concat}, so that the previous representation of a node (i.e., $ \bm{h}_{v}^{k-1} $) can be incorporated at current step $ k $. The intuition behind Eq.~\ref{eq_neib} and Eq.~\ref{eq_concat} is that at each step $ k $, or search depth $ k $, node $ v $ aggregates information from its local neighbors in $ \bm{S}_{v} $. The neighbors also aggregate information following the same procedures. As the process iterates from $ k=1 $ to $ k=K $, node $ v $ gradually gains more and more information from nodes that are in a larger search depth. 

Before generating node representations, the required neighborhood sets (up to search depth $ K $) are sampled for each node in batch $ \bm{B} $. Sample size, $ s_{k} $, determines the number of neighbors sampled at search depth $ k $. By the end of $ K $ steps, the representation of node $ v $ captures the information within its $ K $-hop neighbors. The representation vector at the final step, denoted as $ \bm{h}_{v}^{K} $, serves as the node representation of local view:
\begin{equation}
	\bm{e}_{v}^{A}=\bm{h}_{v}^{K}=f_{A}\left(\bm{x}_{v}, \bm{x}_{S}\right), v\in \bm{B}.
	\label{eq_emb}
\end{equation}
Here, $ f_{A} $ refers to the GNN encoder of local view, and $ \bm{x}_{S} $ is the attribute matrix of the sampled neighbors. 

The node representation of global view is calculated as follows.
\begin{equation}
	\bm{h}_{P}^{k}=\sigma_{1}\left(\bm{W}_{P}^{k}\sum\nolimits_{u\in \bm{P}_{v}}p_{vu}\bm{h}_{u}^{k-1}\right), k\in\left\lbrace 1,\ldots,K \right\rbrace,
	\label{eq_neib_diff}
\end{equation}
where $ p_{vu} $ are diffusion matrix entries in the row corresponding to node $ v $, $ \bm{P}_{v} $ is a set of nodes associated with entries $ p_{vu} $, $ \bm{W}_{P}^{k} $ is the weight matrix, and $ \bm{h}_{P}^{k} $ is the representation vector of node $ v $ at step $ k $. The node representation encoded from global view is
\begin{equation}
	\bm{e}_{v}^{P}=\bm{h}_{P}^{K}=f_{P}\left(\bm{x}_{P}\right), v\in \bm{B},
	\label{eq_emb_diff}
\end{equation}
in which $ f_{P} $ is the GNN encoder of global view and $ \bm{x}_{P} $ is the attribute matrix of neighboring nodes in the augmented graph characterized by diffusion matrix. As suggested by~\cite{klicpera_diffusion_2019}, diffusion matrix, $ \bm{P} $, is sparsified by selecting the highest $ s $ entries per row, which can be interpreted as sampling $ s $ neighboring nodes based on the augmented graph. Note that, since the computation of diffusion matrix already considers a node itself by adding the identity matrix (see Eq.~\ref{eq_trans}), skip connection is removed in the calculation of global-view representation.

Representations of the local and global views are concatenated to obtain the embedding vector of a node:
\begin{equation}
	\bm{e}_{v}=\left[\bm{e}_{v}^{A};\bm{e}_{v}^{P}\right]. 
	\label{eq_emb_concat}
\end{equation}
Node representations are generated with Eq.~\ref{eq_neib}--Eq.~\ref{eq_emb_concat} for both the source and target graphs. The same GNN encoders, $ f_{A} $ and $ f_{P} $, are utilized to compute node representations for each graph. That is, the model parameters are shared when processing these two graphs. The GNN encoders can naturally be trained with a batch of nodes per iteration. Note that both parameter sharing and minibatch training are required by minimax entropy training.
\subsubsection{Contrastive Learning between Structural Views}
\begin{figure}[htbp]
	\centering
	\includegraphics[width=11.00cm]{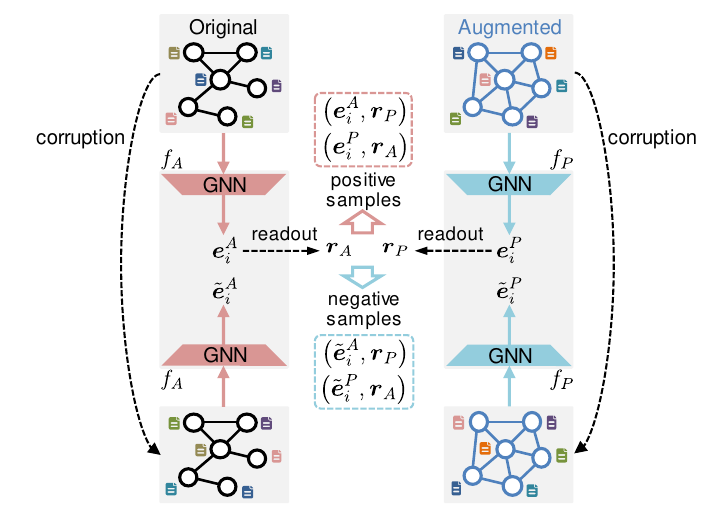}		
	\caption{Contrastive learning between the local view (original graph) and the global view (diffusion-augmented graph).}
	\label{fig_contra_learn}
\end{figure}

As shown in Figure~\ref{fig_contra_learn}, graph contrastive learning is applied to maximize the agreement between representations learned from the local and global views. The GNN encoders are encouraged to simultaneously encode a graph's local and global information. Specifically, we maximize mutual information between the local and global views by contrasting the node representation of one view with a graph-level representation of the other view. 

When considering a batch of nodes, denoted as $ \bm{B} $, GNN encoder $ f_{A} $ computes the local-view representations as $ \bm{E}_{A}=\left[\bm{e}_{1}^{A},\ldots,\bm{e}_{\left| \bm{B} \right|}^{A}\right] $. Readout function, $ \mathcal{R} $, is applied to obtain a summary vector $ \bm{r}_{A} $ by averaging the node representations in this batch:
\begin{equation}
	\bm{r}_{A}=\mathcal{R}\left(\bm{E}_{A}\right) =\sigma_{2}\left(\dfrac{1}{\left| \bm{B}\right|}\sum_{i=1}^{\left| \bm{B} \right|}\bm{e}_{i}^{A}\right).
	\label{eq_read}
\end{equation}
Here, $ \sigma_{2} $ represents the logistic sigmoid activation function, that is, $ \sigma_{2}\left(x\right) =1/\left(1+{\rm exp}\left(-x\right)\right)$. Summary vector, $ \bm{r}_{A} $, serves as a graph-level representation associated with the nodes in this batch and their neighbors considered. Similarly, we can use GNN encoder $ f_{P} $ to obtain the global-view representations $ \bm{E}_{P}=\left[\bm{e}_{1}^{P},\ldots,\bm{e}_{\left| \bm{B} \right|}^{P}\right] $ and the corresponding summary vector $ \bm{r}_{P} $.

The pairs $ \left(\bm{e}_{i}^{A},\bm{r}_{P}\right) $ and $ \left(\bm{e}_{i}^{P},\bm{r}_{A}\right) $ are regarded as positive samples. Mutual information between the local and global views is maximized by classifying the positive samples and their negative counterparts. Following~\cite{velickovic_deep_2019}, row-wise shuffling of attribute matrix $ \bm{X} $ is applied to corrupt the graph. GNN encoders, $ f_{A} $ and $ f_{P} $, take the original adjacency and diffusion matrices, as well as the corrupted attributes, as inputs to calculate the node representations for negative samples, namely $ \tilde{\bm{E}}_{A}=\left[ \tilde{\bm{e}}_{1}^{A},\ldots,\tilde{\bm{e}}_{\left| \bm{B} \right|}^{A}\right] $ and $ \tilde{\bm{E}}_{P}=\left[ \tilde{\bm{e}}_{1}^{P},\ldots,\tilde{\bm{e}}_{\left| \bm{B} \right|}^{P}\right] $. The negative samples are denoted as $ \left(\tilde{\bm{e}}_{i}^{A},\bm{r}_{P}\right) $ and $ \left(\tilde{\bm{e}}_{i}^{P},\bm{r}_{A}\right) $.

To classify the positive and negative samples, we utilize a bilinear function $ \mathcal{D} $ that assigns a score to each sample, indicating its likelihood of being positive. 
\begin{equation}
	\bm{s}_{i}=\mathcal{D}\left(\bm{e}_{i},\bm{r}\right) =\sigma_{2}\left(\bm{e}_{i}^{\top}\bm{W}_{b}\bm{r}\right), i\in\left\lbrace 1,\ldots, \left| \bm{B} \right|\right\rbrace,
	\label{eq_score_pos}
\end{equation}
where $ \bm{W}_{b} $ is a learnable matrix and $ \left(\bm{e}_{i},\bm{r}\right) $ is an example of input sample containing a representation vector $ \bm{e}_{i} $ and a summary vector $ \bm{r} $. Finally, a contrastive loss is defined using the Jensen-Shannon divergence~\cite{liu_graph_2022}:
\begin{equation}
	\begin{aligned}
		\mathcal{L}_{\rm CL} = &-\dfrac{1}{4\left| \bm{B}\right|}\sum_{i=1}^{\left| \bm{B} \right|}\{{\rm log}\mathcal{D}\left(\bm{e}_{i}^{A},\bm{r}_{P}\right) + {\rm log}\mathcal{D}\left(\bm{e}_{i}^{P},\bm{r}_{A}\right) \\&+ {\rm log}\left[ 1-\mathcal{D}\left(\tilde{\bm{e}}_{i}^{A},\bm{r}_{P}\right)\right] + {\rm log}\left[ 1-\mathcal{D}\left(\tilde{\bm{e}}_{i}^{P},\bm{r}_{A}\right)\right]\}.
	\end{aligned}
	\label{eq_mi_cls}
\end{equation}
By optimizing the contrastive loss, the two instances within the positive sample are pulled closer, whereas their counterparts in the negative sample are pushed away. Note that the contrastive loss is independently computed for each of the source and target graphs, namely $ \mathcal{L}_{\rm CL}^{s} $ and $ \mathcal{L}_{\rm CL}^{t} $. Therefore, the overall contrastive loss is reformulated as the sum of the loss on each graph:
\begin{equation}
	\mathcal{L}_{\rm CL} = \mathcal{L}_{\rm CL}^{s} + \mathcal{L}_{\rm CL}^{t}.
	\label{eq_un_loss}
\end{equation}
\subsection{Node Label Prediction}
\label{sec_label_pred}
The cosine similarity-based classifier is found to be effective in the few-shot classification where a few labeled data from the new classes are given to train a classification model~\cite{chen_closer_2019}. The few-shot classification shares certain similarities with the SSDA problem, where a few labeled nodes in the new domain (target domain) are provided. Following~\cite{saito_semi-supervised_2019}, we devise node classifier $ f_{c} $ to be cosine similarity-based:
\begin{equation}
	\hat{\bm{y}}_{v}=f_{c}\left(\bm{e}_{v}\right) =\sigma_{3}\left(\dfrac{1}{T}\dfrac{\bm{W}_{c}^{\top}\bm{e}_{v}}{\left\|  \bm{e}_{v}\right\|_{2}}\right), v\in \bm{B},
	\label{eq_node_cls}
\end{equation}
where $ \bm{W}_{c} $ is the learnable weight matrix; $ T $ is a temperature parameter for scaling; nonlinear activation $ \sigma_{3} $ is a softmax function for multiclass classification; class probability vector $ \hat{\bm{y}}_{v}^{\top} $ is one row in label  prediction matrix $ \hat{\bm{Y}} $. Weight matrix, $ \bm{W}_{c}=\left[\bm{w}_{1},\bm{w}_{2},\ldots,\bm{w}_{C}\right]  $, consists of a series of weight vectors (i.e., $ \bm{w}_{j}, j\in\left\lbrace 1,\ldots,C \right\rbrace $). In order to classify the nodes correctly, the direction of weight vector $ \bm{w}_{j} $ has to be representative to the normalized node representations of class $ j $. Therefore, each weight vector can be treated as an estimated prototype for the corresponding class~\cite{saito_semi-supervised_2019}. 

Cross-entropy loss, $ \mathcal{L}_{\rm CE} $, is computed using the labeled nodes in both the source and target graphs:
\begin{equation}
	\mathcal{L}_{\rm CE}\!=\!-\!\mathop{\mathbb{E}}\nolimits_{v\in \bm{B}^{s}}\!\sum_{j=1}^C \left(Y_{vj}^{s}{\rm log}\hat{Y}_{vj}^{s}\!\right)\!-\!\mathop{\mathbb{E}}\nolimits_{v\in \bm{T}_{l}}\!\sum_{j=1}^C \left(Y_{vj}^{t}{\rm log}\hat{Y}_{vj}^{t}\!\right).
	\label{eq_cross_ent}
\end{equation}
Here, $ \bm{B}^{s} $ represents a batch of source graph nodes, $ \bm{T}_{l} $ denotes the set of labeled nodes in the target graph, binary element $ Y_{vj} $ in label matrix $ \bm{Y} $ indicates if a node $ v $ belongs to class $ j $, and $ \hat{Y}_{vj} $ is the corresponding entry in label prediction matrix $ \hat{\bm{Y}} $. The source and target graphs are distinguished by two superscripts, namely $ s $ and $ t $. By minimizing the cross-entropy loss, the GNN encoders are expected to generate discriminative node representations.  
\subsection{Semi-supervised Domain Adaptation}
\label{sec_ssda}
If the model is optimized using the cross-entropy loss calculated by the labeled source and target nodes, the trained model is likely to be biased towards the source graph, since the source labels are dominant. The node representations of unlabeled target nodes would not be discriminative enough. To address this issue, we apply minimax entropy training~\cite{saito_semi-supervised_2019} as a domain adaptation technique. 

Minimax entropy optimizes the model on the entropy of unlabeled nodes in the target graph. With the model optimized on the cross-entropy loss (Eq.~\ref{eq_cross_ent}), the estimated prototypes (i.e., $ \bm{w}_{j}, j\in\left\lbrace 1,\ldots,C \right\rbrace $) will be closer to the embedding distribution of source graph. Then the ``position" of each prototype $ \bm{w}_{j} $ is moved to the target graph by training the node classifier to increase the entropy of unlabeled target nodes:
\begin{equation}
	\mathcal{L}_{\rm EN}=-\mathop{\mathbb{E}}\nolimits_{v\in\bm{B}^{t}}\sum_{j=1}^C \left(\hat{Y}_{vj}^{t}{\rm log}\hat{Y}_{vj}^{t}\right), 
	\label{eq_entropy}
\end{equation}
where $ \bm{B}^{t} $ is a batch of nodes sampled from the set of unlabeled nodes in target graph, i.e., $ \bm{T}_{u} $. Increasing the entropy leads to a more uniform prediction score for each class, which means each prototype $ \bm{w}_{j} $ shall be similar to all unlabeled target node representations. In contrast, the GNN encoders are optimized to decrease the entropy. The embedding vectors of unlabeled target nodes are expected to be clustered around one of the prototypes by reducing the entropy, which results in discriminative node representations. 

To summarize, the domain adaptation process is modeled as a minimax game between the GNN encoders and the node classifier. Specifically, the node classifier is trained to maximize the entropy, whereas the GNN encoders are optimized to minimize it. We insert a gradient reversal layer~\cite{ganin_domain-adversarial_2016} between the GNN encoders and the node classifier, so that these two model modules can be updated in one backpropagation.

\subsection{Overall Objective and Model Training}
The proposed SemiGCL model is trained with cross-entropy loss $ \mathcal{L}_{\rm CE} $, contrastive loss $ \mathcal{L}_{\rm CL} $, and entropy loss $ \mathcal{L}_{\rm EN} $. The overall learning objective functions are:
\begin{equation}
	\hat{\bm{\theta}}_{g}=\mathop{\rm argmin}\limits_{\bm{\theta}_{g}} \left(\mathcal{L}_{\rm CE}+\lambda_{1}\mathcal{L}_{\rm CL}+\lambda_{2}\mathcal{L}_{\rm EN}\right), 
	\label{eq_FE_loss}
\end{equation}
\begin{equation}
	\hat{\bm{\theta}}_{c}=\mathop{\rm argmin}\limits_{\bm{\theta}_{c}} \left( \mathcal{L}_{\rm CE}-\lambda_{3}\mathcal{L}_{\rm EN}\right), 
	\label{eq_Cls_loss}
\end{equation}
where balance coefficients, $ \lambda_{1} $, $ \lambda_{2} $ and $ \lambda_{3} $, are contrastive learning coefficient, domain adaptation coefficient and entropy coefficient, respectively; $ \bm{\theta}_{g} $ and $ \bm{\theta}_{c} $ are the sets of learnable parameters in the GNN encoders and in the node classifier, respectively.

\begin{algorithm}[h]
	\newcommand{\algrule}[1][.2pt]{\par\vskip.5\baselineskip\hrule height #1\par\vskip.5\baselineskip}
	\caption{SemiGCL}
	\label{algo}	
	\SetKwInOut{Input}{Input}
	\SetKwInOut{Output}{Output}
	\SetKwInOut{Testing}{Testing}	
	\Input{Fully labeled source graph $ \mathcal{G}^{s}\left(\bm{V}^{s},\bm{A}^{s},\bm{X}^{s},\bm{Y}^{s}\right) $; partially labeled target graph $ \mathcal{G}^{t}\left(\bm{V}^{t},\bm{A}^{t},\bm{X}^{t},\bm{Y}^{t}\right) $; batch size; coefficients $ \lambda_{1} $, $ \lambda_{2} $, and $ \lambda_{3} $.}
	\algrule[0.4pt]
	Initialize model parameters $ \bm{\theta}_{g} $ for the GNN encoders and $ \bm{\theta}_{c} $ for the node classifier.\\
	\While{not max epoch}{
		\While{not max iteration}{
			Sample a batch of labeled nodes (i.e., $ \bm{B}^{s} $) from $ \mathcal{G}^{s} $  and a batch of unlabeled nodes (i.e., $ \bm{B}^{t} $) from $ \mathcal{G}^{t} $\;
			Generate node representations in Eq.~\ref{eq_emb_concat}\;
			Calculate contrastive loss $ \mathcal{L}_{\rm CL} $ in Eq.~\ref{eq_un_loss}\;
			Calculate cross-entropy loss $ \mathcal{L}_{\rm CE} $ in Eq.~\ref{eq_cross_ent}\;
			Calculate entropy loss $ \mathcal{L}_{\rm EN} $ in Eq.~\ref{eq_entropy}\;
			Backpropagate and update $ \bm{\theta}_{g} $ and $ \bm{\theta}_{c} $ using the overall losses in Eq.~\ref{eq_FE_loss} and Eq.~\ref{eq_Cls_loss}.\\
		}
	}
	\algrule[0.4pt]   
	\Testing{With model parameters $ \bm{\theta}_{g} $ and $ \bm{\theta}_{c} $ optimized, node representations of unlabeled target nodes are computed using Eq.~\ref{eq_emb_concat}. The corresponding label predictions are subsequently calculated using Eq.~\ref{eq_node_cls}.}
\end{algorithm}

Algorithm~\ref{algo} provides an outline of the main procedures for training and testing. During the training stage, the nodes are initially sampled independently from the source graph and the unlabeled set of the target graph (Line 4). Subsequently, the GNN encoders are utilized to generate representations for both the source and target nodes (Line 5). Following this, the contrastive loss, the cross-entropy loss, and the entropy loss are computed (Lines 6-8). Finally, the trainable parameters in SemiGCL are updated using the overall losses defined in Eq.~\ref{eq_FE_loss} and Eq.~\ref{eq_Cls_loss} (Line 9). With the completion of multiple epochs, the model reaches convergence. The node representations produced by the GNN encoders would become transferrable and discriminative. During the testing stage, we apply the trained node classifier to categorize the unlabeled target nodes using their corresponding node representations.

The time complexity of SemiGCL is derived by investigating its two modules: the GNN encoders (Eq.~\ref{eq_emb} and Eq.~\ref{eq_emb_diff}) and the node classifier (Eq.~\ref{eq_node_cls}). Since the GNN encoders employ neighborhood aggregation, each GNN encoder has time complexity similar to the one of GraphSAGE \cite{hamilton_inductive_2017}, i.e., $ \mathcal{O}\left(\prod_{k=1}^{K}s_{k}\right) $ per node. $ s_{k} $ represents the sample size of neighboring nodes at search depth $ k $. $ K $ is the maximum search depth. The time complexity of node classifier is linear to the number of processed nodes. Consequently, the overall time complexity of SemiGCL is linear with respect to the number of nodes.
\subsection{Theoretical Analysis on Minimax Entropy} 
\label{sec_theory}
Following MME~\cite{saito_semi-supervised_2019}, we theoretically analyze the mechanism of minimax entropy training for reducing domain divergence. As shown in~\cite{ben-david_analysis_2006}, domain divergence can be measured with a domain discriminator. 
Let $f_{d} \in \mathcal{H}$ be a domain discriminator from hypothesis space $ \mathcal{H} $, the ${\cal H}$-divergence between source domain distribution $p$ and target domain distribution $q$ is
\begin{equation}
	\label{eq_h_div}
	{D_\mathcal{H}}(p,q) \triangleq 2\mathop {\sup }\limits_{f_{d}  \in \mathcal{H}} \left| {\mathop {\Pr }\limits_{{\bm{e}} \sim p} \left[ {f_{d} ({\bm{e}}) = 1} \right] - \mathop {\Pr }\limits_{{\bm{e}} \sim q} \left[ {f_{d}({\bm{e}}) = 1} \right]} \right|
\end{equation}
where $ \bm{e} $ denotes the node representation in Eq.~\ref{eq_emb_concat}. This theory reveals that domain divergence can be measured by training a domain discriminator $ f_{d} $ that distinguishes distributions $p$ and $q$.

Although there is no domain discriminator in the model (see Figure~\ref{fig_model_archi}), SemiGCL minimizes the divergence between the source and target graphs through minimax training on the entropy of unlabeled target nodes. Without any additional model modules, the domain discriminator can be regarded as a classifier that assigns a binary domain label for a node representation according to its entropy:
\begin{equation}
	f_{d}(\bm{e}) = 
	\begin{cases}
		1, & \text{if } H(f_{c}(\bm{e})) \geq \gamma,\\
		0, & \text{otherwise},
	\end{cases}
\end{equation}
where $f_{c}$ is the node classifier, $H$ is the entropy, and $\gamma$ is a threshold. As shown in Eq.~\ref{eq_node_cls}, node classifier, $f_{c}$, outputs the probability of class prediction. The ${\cal H}$-divergence in Eq.~\ref{eq_h_div} can be rewritten as 
\begin{equation}
	\label{eq_Hdiverence}
	\begin{aligned}
		{D_\mathcal{H}}(p,q)& \triangleq 2\mathop {\sup }\limits_{f_{d}  \in \mathcal{H}} \left| {\mathop {\Pr }\limits_{{\bm{e}} \sim p} \left[ {f_{d} ({\bm{e}}) = 1} \right] - \mathop {\Pr }\limits_{{\bm{e}} \sim q} \left[ {f_{d}({\bm{e}}) = 1} \right]} \right|\\
		& =\textstyle{ 2\mathop {\sup }\limits_{f_{c}  \in \mathcal{C}}\left| \mathop  {\Pr }\limits_{{\bm{e}} \sim p} \left[ {H(f_{c}(\bm{e})) \geq \gamma} \right] -  \mathop  {\Pr }\limits_{{\bm{e}} \sim q} \left[ {H(f_{c}(\bm{e})) \geq \gamma} \right] \right|}\\
		& \leq  2\mathop {\sup }\limits_{f_{c}  \in \mathcal{C}}  \mathop  {\Pr }\limits_{{\bm{e}} \sim q} \left[ {H(f_{c}(\bm{e})) \geq \gamma} \right],
	\end{aligned}
\end{equation}
where $ \mathcal{C} $ is the hypothesis space of node classifier $f_{c}$. In the final inequality, it is assumed that $\mathop  {\Pr }\limits_{{\bm{e}} \sim p} \left[ {H(f_{c}(\bm{e})) \geq \gamma} \right] \leq  \mathop  {\Pr }\limits_{{\bm{e}} \sim q} \left[ {H(f_{c}(\bm{e})) \geq \gamma} \right]$. The reason is that the entropy of a source node is driven to be a small value by minimizing the cross-entropy loss in Eq.~\ref{eq_cross_ent}.

The inequality in Eq.~\ref{eq_Hdiverence} shows that domain divergence can be bounded by the proportion of target nodes with entropy surpassing threshold $\gamma$. The upper bound can be obtained by finding a node classifier $ f_{c}  \in \mathcal{C} $ that attains maximum entropy for the node representations of target graph. Our goal is to find node representations that achieve the lowest domain divergence. Thus, the minimax objective is 
\begin{equation}
	\min_{{\bm{e}} \sim q}\mathop{\max }\limits_{f_{c} \in \mathcal{C}}  \mathop  {\Pr }\limits_{{\bm{e}} \sim q} \left[ {H(f_{c}(\bm{e})) \geq \gamma} \right].
\end{equation}
It is required to find the GNN encoders that achieve the minimum with respect to node representations ${\bm{e}} \sim q$. The above minimax objective corresponds to the objective functions Eq.~\ref{eq_FE_loss} and Eq.~\ref{eq_Cls_loss}. In summary, the minimax entropy training adopted by SemiGCL can theoretically bound and reduce the domain divergence between the source and target graphs by maximizing entropy and minimizing entropy, respectively. Thus minimax entropy training can align the data distributions of these two graphs, promoting knowledge transfer from the source graph to the target graph.
\section{Experiments}
\label{sec_exp}
In this section, our goal is to address the following research questions (RQs) by conducting extensive experiments.
\begin{itemize}	
	\item RQ1: How does SemiGCL compare with the state-of-the-art baselines in terms of performance on the SSDA tasks?
	\item RQ2: What are the advantages of integrating graph contrastive learning and minimax entropy training in the SemiGCL model?
	\item RQ3: How does the model performance vary depending on the number of labeled nodes per class in the target graph?
	\item RQ4: How do the hyperparameters affect the performance of the SemiGCL model?
\end{itemize}
\subsection{Experimental Setup}
\label{exp_setup}
\subsubsection{Datasets}
\label{sec_data}
Following CDNE~\cite{shen_network_2021}, ACDNE~\cite{shen_adversarial_2020}, and AdaGCN~\cite{dai_graph_2022}, we conduct experiments on five real-world information networks, including three citation networks (i.e., ACMv9, Citationv1, and DBLPv7) and two social networks (i.e., Blog1 and Blog2). Table~\ref{tab_dataset} presents the dataset statistics. Three citation networks are acquired from ArnetMiner~\cite{tang_arnetminer_2008}. These citation networks are represented as undirected graphs where each node corresponds to a paper. An edge between nodes signifies a citation relationship between two papers. The paper title is used to extract the node attribute vector. The union attribute set of these three citation graphs consists of 6,775 node attributes. Each node is assigned to one of the five classes based on the paper's research topic: “Database”, “Artificial Intelligence”, “Computer Vision”, “Information Security”, and “Networking”. 

Blog1 and Blog2 are extracted from the BlogCatalog dataset~\cite{li_unsupervised_2015}. In these social networks, the bloggers and their friendships are represented by nodes and undirected edges, respectively. The attribute vector of each node is derived from the blogger's self-description. Based on the blogger's interest group, each node is assigned to one of the six classes. Unlike the citation networks, Blog1 and Blog2 have larger average degrees, indicating that the nodes in these social networks have more neighbors. 

We assess all methods on eight transfer tasks: C$ \rightarrow $A, D$ \rightarrow $A, A$ \rightarrow $C, D$ \rightarrow $C, A$ \rightarrow $D, C$ \rightarrow $D, B2$ \rightarrow $B1, and B1$ \rightarrow $B2. ACMv9, Citationv1, DBLPv7, Blog1, and Blog2 are represented by A, C, D, B1, and B2, respectively. The arrow symbol, ``$ \rightarrow $", shows the transfer of knowledge from a source graph to a target graph. Table~\ref{tab_R_a} presents the common attribute rate of each transfer task, which indicates the discrepancy between the attribute distributions of the source and target graphs. Since the two social networks originate from the BlogCatalog dataset, their attribute distributions are relatively close.

\begin{table}[htbp]
	\caption{Summary of datasets.}
	\label{tab_dataset}
	\centering
	\centerline{
		\begin{threeparttable}
			\begin{tabular}{l|c|c|c|c|c|c}
				\toprule
				Dataset    & \#Nodes & \#Edges & Average Degree & \#Attributes &  \#Union Attributes  &     \#Classes     \\ \midrule
				ACMv9      &  8,661  & 13,590  &      3.13      &    5,571     & \multirow{3}*{6,775} & \multirow{3}*{5} \\
				Citationv1 &  8,724  & 14,798  &      3.39      &    5,379     &                      &                  \\
				DBLPv7     &  5,463  &  8,098  &      2.96      &    4,412     &                      &                  \\ \midrule
				Blog1      &  2,300  & 33,471  &     29.11      &    4,121     & \multirow{2}*{4,185} & \multirow{2}*{6} \\
				Blog2      &  2,896  & 53,836  &     37.18      &    4,158     &                      &                  \\ \bottomrule
			\end{tabular}
			\begin{tablenotes}
				\footnotesize
				\item[$ \ast $] ``\#Nodes" means the number of nodes.
			\end{tablenotes}
		\end{threeparttable}}
\end{table}
\begin{table}[htbp]
	\caption{Common attribute rate on each transfer task.}
	\label{tab_R_a}
	\centering
	\centerline{
	\begin{tabular}{l|cc|cc|cc|cc}
		\toprule
		Transfer Task         & C$ \rightarrow $A & A$ \rightarrow $C & D$ \rightarrow $A & A$ \rightarrow $D & D$ \rightarrow $C & C$ \rightarrow $D & B2$ \rightarrow $B1 & B1$ \rightarrow $B2 \\ \midrule
		\#Common Attributes   &      \multicolumn{2}{c|}{4,285}       &      \multicolumn{2}{c|}{3,621}       &      \multicolumn{2}{c|}{3,783}       &         \multicolumn{2}{c}{4,094}         \\
		\#Union Attributes    &      \multicolumn{2}{c|}{6,665}       &      \multicolumn{2}{c|}{6,362}       &      \multicolumn{2}{c|}{6,008}       &         \multicolumn{2}{c}{4,185}         \\
		Common Attribute Rate &     \multicolumn{2}{c|}{64.29\%}      &     \multicolumn{2}{c|}{56.92\%}      &     \multicolumn{2}{c|}{62.97\%}      &        \multicolumn{2}{c}{97.83\%}        \\ \bottomrule
	\end{tabular}}
\end{table}
\subsubsection{Baselines} 
Following~\cite{shen_network_2021, wu_unsupervised_2020, shen_adversarial_2020, dai_graph_2022}, the baseline approaches are of three categories: (I) classical GNN models for representation learning on a single graph, (II) typical domain adaptation approaches, and (III) state-of-the-art models designed for cross-graph node classification. The implementation details of our method and the baselines are provided in~\ref{appen_imple_details}.

\begin{enumerate}[(I)]
	\item GCN~\cite{kipf_semi-supervised_2017}, GAT~\cite{velickovic_graph_2018}, GraphSAGE~\cite{hamilton_inductive_2017}, DGI~\cite{velickovic_deep_2019}, and MVGRL~\cite{hassani_contrastive_2020}: They are popular GNN models for single-graph representation learning. GCN designs a simplified filter to conduct spectral graph convolution. GAT specifies learnable weights to the neighboring nodes when aggregating their information. Rather than utilizing the complete neighborhood, GraphSAGE produces the representation of a node by aggregating information from a sampled set of neighbors. DGI and MVGRL are GNN models that devise contrastive losses to maximize the mutual information between a graph's local and global representations. These GNN models are adapted and evaluated under the cross-graph scenario to show whether the classical single-graph GNN models are sufficient for cross-graph node classification tasks.
	\item DANN~\cite{ganin_domain-adversarial_2016}, WDGRL~\cite{shen_wasserstein_2018}, and MME~\cite{saito_semi-supervised_2019}: They are typical domain adaptation approaches with the assumption that the input data are independent and identically distributed, such as images and text. These approaches are evaluated to show whether they can be directly applied to conduct domain adaptation on graphs where the nodes are connected by edges, violating the i.i.d. assumption. To process graph data, they are implemented to only take node attributes as input, ignoring the edge connections.
	\item CDNE~\cite{shen_network_2021}, ACDNE~\cite{shen_adversarial_2020}, AdaGCN~\cite{dai_graph_2022}, UDA-GCN~\cite{wu_unsupervised_2020}, ASN~\cite{zhang_adversarial_2021}, AdaGIn~\cite{xiao_domain_2022}, MFRReg~\cite{you_graph_2023}, and GRADE~\cite{wu_non-iid_2023}: They are state-of-the-art models proposed for cross-graph node classification. These methods are hereafter referred to as cross-graph models. CDNE extracts node representations with stacked autoencoders and incorporates the MMD loss~\cite{long_transfer_2013} to align graph distributions. ACDNE captures topology proximity and attribute affinity with two independent feature extractors. UDA-GCN utilizes the dual graph convolutional network. The gradient reversal layer~\cite{ganin_domain-adversarial_2016} is implemented in ACDNE and UDA-GCN to reduce domain discrepancy. AdaGCN adopts GCN \cite{kipf_semi-supervised_2017} as the feature extractor and alleviates domain divergence by reducing the Wasserstein distance~\cite{shen_wasserstein_2018}. ASN separates the features shared across domains from the domain-private features. AdaGIn is a GNN-based model that matches the multimodal embedding distributions with conditional adversarial networks. MFRReg improves the transferability of UDA-GCN with spectral regularization. GRADE reduces the graph subtree discrepancy to align the graph data distributions.
\end{enumerate}
\subsection{Performance Study (RQ1)}
\label{sec_per_sty}
In this section, the proposed method and the baselines are evaluated on the cross-graph node classification tasks. Under this cross-graph scenario, all models perform node classification tasks by leveraging information from both a fully labeled source graph and a partially labeled target graph. We randomly select five nodes per class (i.e., $ n=5 $) to form the labeled set of target graph. Classification accuracy is reported to quantify the performance of a model in classifying the unlabeled nodes of target graph. As a supplementary evaluation, we also conduct semi-supervised node classification on the partially labeled target graph. In such a single-graph scenario, the models are trained and evaluated with the target graph only.
\subsubsection{Cross-graph Node Classification on Citation Graphs}
\label{sec_cross_cls_cita}
The first category of baselines includes the classical GNN models proposed for single-graph representation learning, i.e., GCN, GAT, GraphSAGE, DGI, and MVGRL. As introduced in~\ref{appen_imple_details}, these GNN models are adapted to conduct cross-graph node classification. In Table~\ref{tab_perf_cross_cita}, with the help of contrastive learning, MVGRL and DGI outperform other GNN models in this category. However, due to the lack of capability to address the domain gap, the accuracies of these GNN models are lower than the cross-graph models in the third category of baselines.

In the second category of baselines (i.e., DANN, WDGRL, and MME), domain adaptation techniques are applied to reduce the domain discrepancy. However, the performance of these approaches is still inferior compared with the models in other categories. This reveals that the node representations generated in citation graphs lack discriminative power when only node attributes are utilized. Moreover, as the attribute distributions are distinct (refer to Table~\ref{tab_R_a}), incorporating the graph structure is necessary. Otherwise, the domain gap would remain significantly large and cannot be reduced effectively using these domain adaptation techniques.

In Table~\ref{tab_perf_cross_cita}, SemiGCL ranks first on all six transfer tasks. For example, SemiGCL outperforms the best-performing baseline (i.e., MFRReg) by 2.12\% on task D$ \rightarrow $C and 1.44\% on task D$ \rightarrow $A. On average, the performance gain of SemiGCL over MFRReg is 1.06\% on the citation graphs. It demonstrates the superiority of SemiGCL under the SSDA setting by incorporating graph contrastive learning and minimax entropy training in a principled way. Although CDNE and AdaGCN also investigate the adaptation scenario where the target graph is partially labeled, their performance on the SSDA tasks is even inferior to some cross-graph models designed for the UDA setting, including ACDNE, UDA-GCN, AdaGIn, and MFRReg. As introduced in~\ref{appen_imple_details}, the UDA cross-graph models are extended to the SSDA setting by optimizing the models with the cross-entropy loss of labeled source and target nodes. However, the learned models would be biased towards the source graph, since the labeled source nodes are of a much greater amount. SemiGCL adopts the minimax entropy training to address this issue and reduce the domain divergence. In addition, the GNN encoders naturally exploit both graph structure and node attributes to generate node representations. Graph contrastive learning maximizes the mutual information between representations encoded from the local and global views of a graph. By doing so, the GNN encoders are encouraged to capture the local and global information of a graph, thereby producing more discriminative node representations to promote node classification.
\begin{table}[htbp]
    \setlength\tabcolsep{2pt}
	\caption{Accuracy (\%) of node classification on target citation graph.}
	\label{tab_perf_cross_cita}
	\centering
	\centerline{
	\begin{threeparttable}
		\begin{tabular}{l|cccccc|c}
			\toprule
			Method                                    &     C$ \rightarrow $A      &     D$ \rightarrow $A      &     A$ \rightarrow $C      &     D$ \rightarrow $C      &     A$ \rightarrow $D      &     C$ \rightarrow $D      &          Average           \\ \midrule
			GCN~\cite{kipf_semi-supervised_2017}      &           72.93 (0.24)           &           68.73 (0.78)           &           77.57 (0.20)           &           74.55 (0.48)           &           72.79 (0.47)           &           74.60 (0.36)           &           73.53            \\
			GAT~\cite{velickovic_graph_2018}          &           71.32 (1.73)           &           67.60 (1.99)           &           77.99 (0.15)           &           75.43 (0.63)           &           73.55 (0.65)           &           74.95 (0.24)           &           73.47            \\
			GraphSAGE~\cite{hamilton_inductive_2017}  &           65.55 (1.09)           &           59.28 (1.76)           &           69.22 (1.66)           &           64.50 (1.48)           &           64.73 (1.60)           &           68.91 (0.46)           &           65.37            \\
			DGI~\cite{velickovic_deep_2019}           &           73.05 (0.49)           &           70.95 (0.44)            &           78.08 (0.57)           &           76.07 (0.34)           &           72.26 (0.38)           &           73.66 (0.56)           &           74.01            \\
			MVGRL~\cite{hassani_contrastive_2020}     &           73.15 (0.45)           &           70.01 (0.36)           &           77.77 (0.54)           &           76.49 (0.42)           &           74.07 (0.25)           &           76.54 (0.42)           &           74.67            \\ \midrule
			DANN~\cite{ganin_domain-adversarial_2016} &           52.87 (0.31)           &           49.72 (0.68)           &           55.77 (0.50)           &           54.71 (0.60)           &           56.95 (1.26)           &           56.80 (0.49)           &           54.47            \\
			WDGRL~\cite{shen_wasserstein_2018}        &           52.33 (0.16)           &           49.38 (0.42)           &           55.46 (0.31)           &           53.52 (0.57)           &           56.35 (1.51)           &           56.71 (0.57)           &           53.96            \\
			MME~\cite{saito_semi-supervised_2019}     &           52.12 (0.52)           &           49.03 (0.47)           &           55.30 (0.26)           &           52.94 (0.60)           &           56.26 (0.22)           &           57.68 (0.25)           &           53.89            \\ \midrule
			CDNE~\cite{shen_network_2021}             &           76.55 (0.35)           &           73.23 (0.53)           &           80.00 (0.25)           &           78.75 (0.66)           &           74.76 (0.26)           &           74.42 (0.19)           &           76.29            \\
			ACDNE~\cite{shen_adversarial_2020}        &     77.97 (0.21)      &     74.44 (0.33)      &     \underline{83.70} (0.12)     &     81.93 (0.40)     &     77.68 (0.13)     &     78.01 (0.30)     &     78.96      \\
			AdaGCN~\cite{dai_graph_2022}              &           76.45 (0.50)           &           74.29 (0.40)           &           82.16 (0.45)           &           80.49 (0.30)           &           76.75 (0.34)           &           77.17 (0.31)           &           77.89            \\
			UDA-GCN~\cite{wu_unsupervised_2020}       &           77.48 (0.54)           &           75.00 (0.55)           &           82.06 (0.15)           &           80.43 (0.45)           &           77.45 (0.24)           &           78.55 (0.42)            &           78.50            \\ 
			ASN~\cite{zhang_adversarial_2021}       &           77.86 (0.74)           &           75.19 (0.99)           &           81.05 (0.82)           &           80.06 (0.92)           &           75.61 (0.68)           &           75.58 (1.37)           &           77.56            \\
			AdaGIn~\cite{xiao_domain_2022}       &           77.08 (0.36)           &           75.86 (0.72)           &           83.17 (0.21)           &           \underline{82.92} (0.29)           &           76.82 (0.45)           &           77.03 (0.35)           &           78.81            \\
   			MFRReg~\cite{you_graph_2023}       &           \underline{78.03} (0.45)           &           \underline{76.32} (0.75)           &           82.78 (0.15)           &          81.73 (0.29)           &           \underline{77.93} (0.29)           &           \underline{78.78} (0.35)           &           \underline{79.26}            \\
      		GRADE~\cite{wu_non-iid_2023}       &           74.01 (0.52)           &           70.00 (0.37)           &           79.04 (0.31)           &           76.20 (0.60)           &           74.23 (0.42)           &           76.30 (0.38)           &           74.96            \\\midrule
			SemiGCL [ours]                            & \textbf{79.03} (0.38) & \textbf{77.76} (0.56) & \textbf{83.89} (0.20) & \textbf{83.85} (0.38) & \textbf{78.00} (0.60) & \textbf{79.36} (0.49) & \textbf{80.32} \\ \bottomrule
		\end{tabular}
		\begin{tablenotes}\footnotesize
			\item[*] A: ACMv9, C: Citationv1, D: DBLPv7. In each column, the highest classification accuracy is in boldface, and the second-best accuracy is underlined. The values in parentheses are standard deviations.
		\end{tablenotes}
	\end{threeparttable}}
\end{table} 
\subsubsection{Cross-graph Node Classification on Social Graphs}
\label{sec_cross_cls_social}
In Table~\ref{tab_perf_cross_social}, SemiGCL performs best on transfer tasks B2$ \rightarrow $B1 and B1$ \rightarrow $B2. Compared with the best-performing baseline (i.e., ACDNE), the performance gain yielded by SemiGCL is 1.08\% on average. The accuracies of SemiGCL on the social graphs are higher than its results on the citation graphs. Blog1 and Blog2 have closer attribute distributions, which is supported by the largest common attribute rate shown in Table~\ref{tab_R_a}. Therefore, the domain divergence between Blog1 and Blog2 is not as large as the one between two citation graphs, reducing the difficulty of transfer tasks. Note that, in this case, the classical domain adaptation methods (i.e., DANN, WDGRL, and MME) also yield impressive performance.
\begin{table}[htbp]
	\setlength\tabcolsep{1pt}
	\caption{Accuracy (\%) of node classification on target social graph (Categories I and II).}
	\label{tab_perf_cross_social-I-II}
	\centering
	\scalebox{0.9}{
		\centerline{
			\begin{threeparttable}
				\begin{tabular}{c|ccccc|ccc}
					\hline
					Method        & GCN & GAT & GraphSAGE & DGI & MVGRL & DANN & WDGRL & MME        \\ \hline
					B2$ \rightarrow $B1 &                74.61 (1.29)                 &              66.33 (0.80)               &                  87.68 (0.24)                  &              70.42 (0.30)             &                 78.03 (0.73)                &                   86.81 (0.25)                   &               86.31 (0.28)                &                 87.67 (0.16) \\
					B1$ \rightarrow $B2 &                74.24 (0.63)                 &              67.33 (1.41)              &                  86.62 (0.47)                  &              71.65 (0.37)             &                 79.88 (0.61)                &                   86.13 (0.33)                   &               85.64 (0.33)               &                 86.76 (0.55) \\ \hline
					Average       &                74.43                 &              66.83               &                  87.15                   &              71.04              &                 78.96                 &                   86.47                   &               85.98                &                 87.22 \\ \hline
				\end{tabular}
				\begin{tablenotes}\normalsize
					\item[*] B1: Blog1, B2: Blog2. The values in parentheses are standard deviations.
				\end{tablenotes}
	\end{threeparttable}}}	
\end{table}
\begin{table}[htbp]
	\setlength\tabcolsep{1pt}
	\caption{Accuracy (\%) of node classification on target social graph (Category III).}
	\label{tab_perf_cross_social}
	\centering
	\scalebox{0.9}{
		\centerline{
			\begin{threeparttable}
				\begin{tabular}{c|cccccccc|c}
					\hline
					Method        & CDNE & ACDNE & AdaGCN & UDA-GCN & ASN & AdaGIn & MFRReg & GRADE &       SemiGCL       \\ \hline
					B2$ \rightarrow $B1 &             81.79 (0.77)             &         \underline{91.79} (0.68)         &            75.95 (0.75)            &                69.28 (1.81) & 68.75 (2.38) & 91.67 (0.12) & 70.43 (1.83) & 74.67 (0.51)           & \textbf{92.94} (0.21) \\
					B1$ \rightarrow $B2 &             84.96 (0.72)            &         \underline{91.93} (0.11)         &            75.12 (0.47)            &                70.28 (1.15) & 68.90 (1.33)  & 90.98 (0.19) & 70.80 (0.88) & 74.61 (0.46)            & \textbf{92.94} (0.23) \\ \hline
					Average       &             83.38             &         \underline{91.86}          &            75.54             &                69.78 & 68.83 & 91.33 & 70.62 & 74.64              & \textbf{92.94} \\ \hline
				\end{tabular}
				\begin{tablenotes}\normalsize
					\item[*] B1: Blog1, B2: Blog2. In each row, the highest classification accuracy is in boldface, and the second-best accuracy is underlined. The values in parentheses are standard deviations.
				\end{tablenotes}
	\end{threeparttable}}}	
\end{table}

\begin{figure}[htbp]
	\makebox[\textwidth][c]{\includegraphics[width=12.8cm]{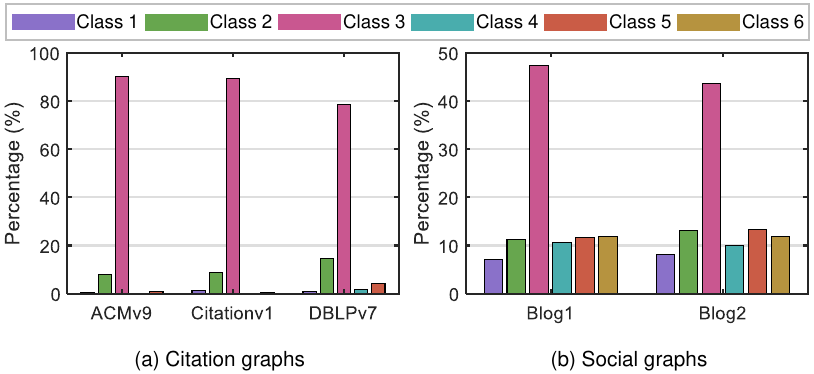}}	
	\caption{Node class distribution in the neighborhood of a node that belongs to Class 3. On every graph, the reported percentage of each class is obtained by averaging the statistics of all nodes that are of Class 3.}
	\label{fig_neib_analysis}
\end{figure}

GNN models commonly adopt smoothing operations to produce similar representations for the nodes with edge connections, possibly leading to the same class prediction for the connected nodes on the subsequent classification task~\cite{li_deeper_2018}. The smoothing operations rely on the homophily assumption that the nodes connected by an edge tend to be of the same class~\cite{kipf_semi-supervised_2017, ma_is_2022}. In Figure~\ref{fig_neib_analysis}, we select nodes of Class 3, referred to as central nodes for clarity, to analyze the class information of neighboring nodes. In the citation graphs, the majority of neighbors belong to the same class as the central node. However, in the social graphs, less than 50\% of the neighboring nodes share the same class with the central node. It means more neighboring nodes violate the homophily assumption in the social graphs. Hence, the neighborhood information around a node is noisier in the social graphs.

It can be seen from Table~\ref{tab_perf_cross_social-I-II} and Table~\ref{tab_perf_cross_social} that the noise within neighborhood results in the clear underperformance of some GNN models, including GCN, GAT, DGI, MVGRL, AdaGCN, UDA-GCN, ASN, MFRReg, and GRADE. The reason is that these GNN models consider the complete neighborhood of a node to compute node representation. In contrast, other GNN models (i.e., GraphSAGE, AdaGIn, and SemiGCL) improve performance by sampling a few neighbors to generate node representations, since neighborhood sampling introduces fewer neighbors that are not of the same class as the central node.
\begin{figure}[htbp]
	\centering
	\includegraphics[width=12.0cm]{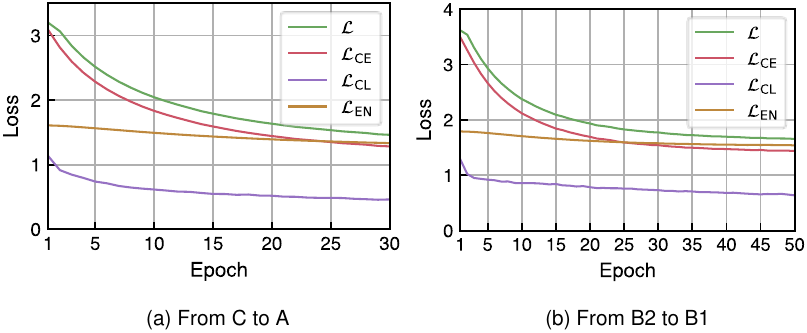}	
	\caption{Loss over training epoch.}
	\label{fig_loss_curve}
\end{figure}
\subsubsection{Loss Evolution during Model Training}
Figure~\ref{fig_loss_curve} demonstrates the evolution of loss values as the GNN encoders are trained over multiple epochs. Two transfer tasks, C$ \rightarrow $A and B2$ \rightarrow $B1, serve as examples. Referring to Eq.~\ref{eq_FE_loss}, the GNN encoders are optimized using an overall loss function $ \mathcal{L}=\mathcal{L}_{\rm CE}+\lambda_{1}\mathcal{L}_{\rm CL}+\lambda_{2}\mathcal{L}_{\rm EN} $. The overall loss consistently decreases with a slower descent rate, revealing that the model gradually converges during training. Similar decreasing trends can be observed in cross-entropy loss $ \mathcal{L}_{\rm CE} $, contrastive loss $ \mathcal{L}_{\rm CL} $, and entropy loss $ \mathcal{L}_{\rm EN} $.
\subsubsection{Single-graph Node Classification}
\label{sec_sin_cls}
Since the target graph has five labeled nodes per class (i.e., $ n=5 $), we can train the node classification models by solely utilizing the available information in target graph. Such a scenario is referred to as the single-graph node classification, as the source graph is not involved in the model training. The trained models are evaluated on the unlabeled nodes of target graph. Although this work focuses on the cross-graph node classification problem under the SSDA setting, we report classification accuracies under the single-graph scenario in Table~\ref{tab_perf_sin} as a supplementary evaluation. Note that some baselines in Table~\ref{tab_perf_cross_cita} are not applicable in the single-graph scenario, including the domain adaptation approaches (Category II) and the cross-graph models (Category III). To conduct node classification in the single-graph scenario, we adapt SemiGCL by optimizing the model without entropy loss $ \mathcal{L}_{\rm EN} $ (see Eq.~\ref{eq_FE_loss} and Eq.~\ref{eq_Cls_loss}).

In Table~\ref{tab_perf_sin}, SemiGCL outperforms the classical GNN models by large margins. On average, the improvements of SemiGCL over MVGRL are 2.19\% on the citation graphs and 6.23\% on the social graphs. Therefore, SemiGCL demonstrates its strength in generating discriminative node representations for a graph. SemiGCL has an average accuracy of 74.36\% on the citation graphs and 72.58\% on the social graphs, which are much lower than its results on the cross-graph node classification, i.e., 80.32\% (Table~\ref{tab_perf_cross_cita}) and 92.94\% (Table~\ref{tab_perf_cross_social}), respectively. As an example, we analyze the results with ACMv9 chosen as the target graph. SemiGCL has an accuracy of 79.03\% on transfer task C$ \rightarrow $A, exceeding its result on the single-graph scenario (i.e., 70.91\%) by 8.12\%. These observations motivate us to transfer knowledge from a labeled source graph, so that the classification performance on target graph can be improved. Note that, MVGRL and DGI also incorporate contrastive learning, resulting in higher accuracies than those of GCN, GAT, and GraphSAGE.
\begin{table}[htbp]
	\caption{Accuracy (\%) of single-graph node classification.}
	\label{tab_perf_sin}
	\centering
	\centerline{
	\begin{threeparttable}
		\begin{tabular}{c|c|c|c|c|c|c}
			\toprule
			Method  & GCN~\cite{kipf_semi-supervised_2017} & GAT~\cite{velickovic_graph_2018} & GraphSAGE~\cite{hamilton_inductive_2017} & DGI~\cite{velickovic_deep_2019} & MVGRL~\cite{hassani_contrastive_2020} &       SemiGCL [ours]       \\ \midrule
			A    &                48.17                 &              48.91               &                  43.24                   &        \underline{70.42}        &                 68.34                 & \textbf{70.91} \\
			C    &                58.94                 &              59.07               &                  49.24                   &              71.95              &           \underline{75.99}           & \textbf{79.66} \\
			D    &                53.57                 &              54.37               &                  49.28                   &              67.50              &           \underline{72.18}           & \textbf{72.50} \\
			Average &                53.56                 &              54.12               &                  47.25                   &              69.96              &           \underline{72.17}           & \textbf{74.36} \\ \midrule\midrule
			B1    &                61.53                 &              55.37               &                  60.55                   &              62.53              &           \underline{67.81}           & \textbf{72.14} \\
			B2    &                61.28                 &              57.31               &                  62.69                   &              61.61              &           \underline{64.88}           & \textbf{73.02} \\
			Average &                61.41                 &              56.34               &                  61.62                   &              62.07              &           \underline{66.35}           & \textbf{72.58} \\ \bottomrule
		\end{tabular}
		\begin{tablenotes}\footnotesize
			\item[*] A: ACMv9, C: Citationv1, D: DBLPv7, B1: Blog1, B2: Blog2. In each row, the highest classification accuracy is in boldface, and the second-best accuracy is underlined.
		\end{tablenotes}
	\end{threeparttable}}
\end{table}
\subsection{Ablation Study (RQ2)}
\label{sec_abla_sty}
In this section, we study four key components in the SemiGCL model, including contrastive learning (CL), global view (GV), local view (LV), and domain adaptation (DA). Each component is removed from the SemiGCL model individually to investigate its contribution. We construct the following model variants.
\begin{itemize}
	\item SemiGCL-CL: A variant of SemiGCL without contrastive learning (CL). The contrastive loss is removed from the overall objective function (i.e., Eq.~\ref{eq_FE_loss}).
	\item SemiGCL-GV: A variant of SemiGCL without learning node representations from the global view (GV). The node embedding vector in Eq.~\ref{eq_emb_concat} is the representation of local view.
	\item SemiGCL-LV: A variant of SemiGCL without learning node representations from the local view (LV). The embedding vector of a node (see Eq.~\ref{eq_emb_concat}) is the representation of global view.
	\item SemiGCL-DA: A variant of SemiGCL without domain adaptation (DA). We remove the entropy loss from the objective functions Eq.~\ref{eq_FE_loss} and Eq.~\ref{eq_Cls_loss}.
\end{itemize}
Note that, in SemiGCL-GV and SemiGCL-LV, as the embedding vector is only extracted from one structural view (i.e., local view or global view), the contrastive loss (Eq.~\ref{eq_un_loss}) cannot be calculated to optimize the model. In this section, the labeled set of target graph is identical to that of Section~\ref{sec_per_sty}. Each class in the target graph has five labeled nodes, i.e., $ n=5 $.
\subsubsection{Performance of Model Variants}
\label{sec_model_var}
Table~\ref{tab_abla_acc} shows that the removal of every component leads to performance drops on all transfer tasks. It reveals that each component contributes to improving the performance of SemiGCL. In particular, with domain adaptation applied, the averaged classification accuracy increases by 1.29\% on citation graphs and 1.21\% on social graphs. Therefore, the model performance can be consistently improved by minimax entropy training. 
\begin{table}[htbp]
	\caption{Accuracy (\%) of SemiGCL variants on the target graph.}
	\label{tab_abla_acc}
	\centering
	\centerline{
	\begin{threeparttable}
		\begin{tabular}{c|c|c|c|c|c}
			\toprule
			Model Variant    & SemiGCL & SemiGCL-CL & SemiGCL-GV & SemiGCL-LV & SemiGCL-DA \\ \midrule
			C$ \rightarrow $A  &  \textbf{79.03}  &   74.81    &   67.26    &   74.45    &   78.29    \\
			D$ \rightarrow $A  &  \textbf{77.76}  &   70.04    &   63.54    &   71.04    &   75.42    \\
			A$ \rightarrow $C  &  \textbf{83.89}  &   80.43    &   74.05    &   79.02    &   82.82    \\
			D$ \rightarrow $C  &  \textbf{83.85}  &   78.96    &   72.10    &   77.02    &   82.41    \\
			A$ \rightarrow $D  &  \textbf{78.00}  &   74.05    &   68.68    &   72.34    &   76.73    \\
			C$ \rightarrow $D  &  \textbf{79.36}  &   76.25    &   70.16    &   75.58    &   78.53    \\
			Average       &  \textbf{80.32}  &   75.76    &   69.30    &   74.91    &   79.03    \\ \midrule\midrule
			B2$ \rightarrow $B1 &  \textbf{92.94}  &   92.56    &   89.76    &   88.38    &   91.74    \\
			B1$ \rightarrow $B2 &  \textbf{92.94}  &   92.10    &   89.12    &   89.14    &   91.72    \\
			Average       &  \textbf{92.94}  &   92.33    &   89.44    &   88.76    &   91.73    \\ \bottomrule
		\end{tabular}
		\begin{tablenotes}
			\footnotesize
			\item[$ \ast $] CL: contrastive learning, GV: global view, LV: local view, DA: domain adaptation. The sign ``-" indicates the removal of one component. The highest accuracy in each row is in boldface.
		\end{tablenotes}	    	
	\end{threeparttable}}
\end{table}

On average, contrasting the local and global views increases the classification accuracy by 4.56\% on the citation graphs and 0.61\% on the social graphs. In comparison with the citation graphs, the social graphs have fewer nodes and larger average degrees (see Table~\ref{tab_dataset}). In a small-scale and dense social graph, the nodes would be “closer”. In other words, it takes a short random walk to reach another node from one node. Therefore, if diffusion is applied to a social graph, the established edges in the augmented graph are likely to be between nodes that appear together within short random walks in the original graph. The information of these nodes could be accessed in the neighborhood aggregation process based on the adjacency matrix (i.e., local view). In such case, the local-view and global-view representations would already have a certain degree of agreement, which makes contrastive learning not that powerful on the social graphs.

In the citation graphs, it is observed that the removal of global-view representation leads to an 11.02\% decrease in average accuracy, which is much more significant than the corresponding decrease in the social graphs (i.e., 3.50\%). Compared with the social graphs, a citation graph is sparser and in a larger scale (see Table~\ref{tab_dataset}). It is more difficult to capture the global information of a citation graph with the local-view neighborhood aggregation. In this case, the global information extracted from the global view is essential for representation learning.  
\subsubsection{Visualization of Node Representations}
\begin{figure}[htbp]
	\centering
	\includegraphics[width=10.6cm]{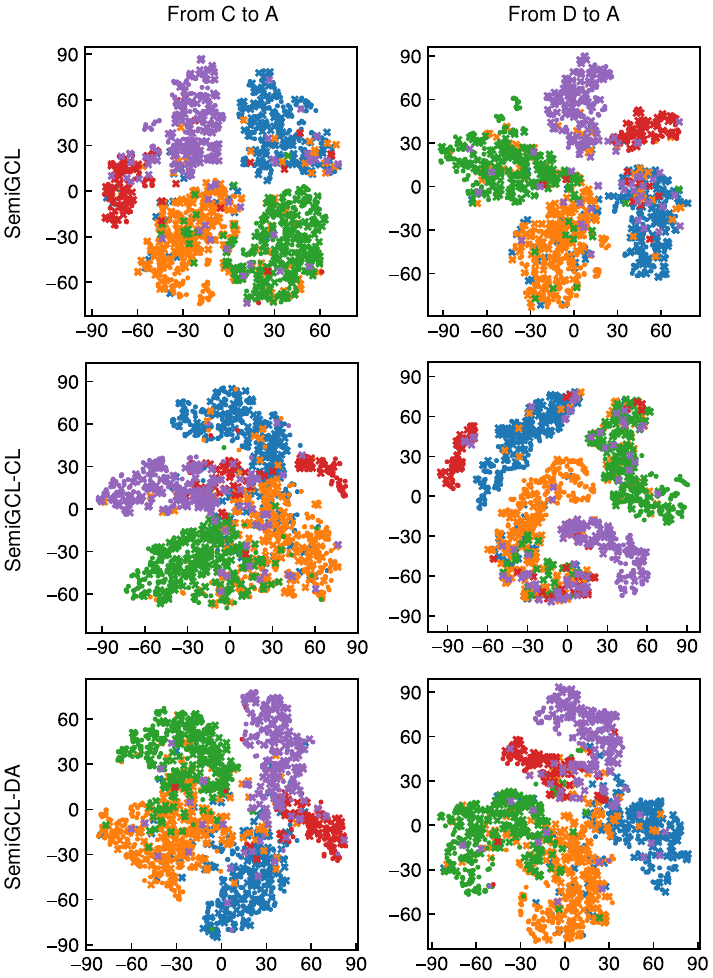}	
	\caption{Node representation visualization with t-SNE. Circles and crosses represent the source and target graph nodes, respectively. Five colors distinguish different paper classes.}
	\label{fig_tsne_visual}
\end{figure}
In Figure~\ref{fig_tsne_visual}, we plot the t-SNE~\cite{maaten_visualizing_2008} visualization of node representations produced by the SemiGCL variants. Two transfer tasks, C$ \rightarrow $A and D$ \rightarrow $A, are taken as examples, with ACMv9 being the target graph. Node representations produced by SemiGCL exhibit the most favorable clustering structure, with a clearer separation between the clusters of different classes. Additionally, node representations from the source and target graphs are mostly close if they belong to the same class. The t-SNE visualization supports that SemiGCL can achieve both the discriminability of node representations and domain alignment. With the contrastive learning removed (see SemiGCL-CL), the nodes of different classes are more likely to be mixed. It reveals that contrastive learning can improve the discriminability of node representations. Similar observations are found in the case without domain adaptation (see SemiGCL-DA).
\subsection{Effect of Target Labeled Number (RQ3)}
In this section, we investigate the model performance when the number of labeled nodes per class (i.e., $ n $) varies from 0 to 10 in the target graph, that is, $ n\in\left\lbrace 0,5, 10\right\rbrace$. In the case that $ n=0 $, the target graph is completely unlabeled, which corresponds to the UDA setting. When there are five labeled nodes per class (i.e., $ n=5 $), the labeled set of target graph is the same as the one in Section~\ref{sec_per_sty}. As introduced in Section~\ref{sec_rel_work}, CDNE and AdaGCN are cross-graph models that investigate the adaptation scenario with a partially labeled target graph. In addition to CDNE and AdaGCN, we also make comparisons with three competitive UDA cross-graph models, i.e., ACDNE, AdaGIn, and MFRReg. 

\begin{figure}[htbp]
	\centering
	\includegraphics[width=10.16cm]{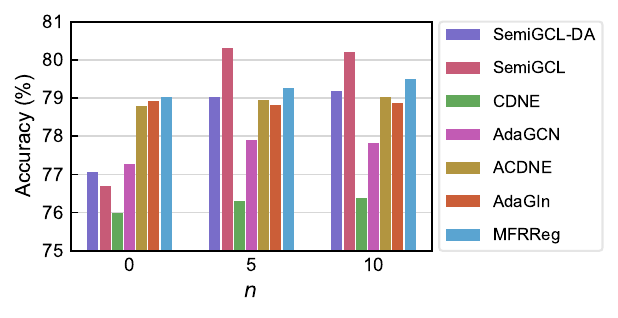}		
	\caption{Average accuracy of the six transfer tasks on the citation graphs. Three numbers of labeled target nodes per class (0, 5 and 10) are tested for comparison among six models.}
	\label{fig_label_num}
\end{figure}

Figure~\ref{fig_label_num} shows the average accuracy of each model on the six transfer tasks of citation graphs, i.e., C$ \rightarrow $A, D$ \rightarrow $A, A$ \rightarrow $C, D$ \rightarrow $C, A$ \rightarrow $D, and C$ \rightarrow $D. The SemiGCL variant without domain adaptation (i.e., SemiGCL-DA) is also included. SemiGCL is the top-performing model on the SSDA tasks, i.e., $ n=5 $ and $ n=10 $. A significant performance lift is observed in SemiGCL from $ n=0 $ to $ n=5 $. Therefore, SemiGCL can effectively exploit a few labeled nodes in the target graph to improve the model performance. In contrast, the accuracies of UDA approaches (i.e., ACDNE, AdaGIn, and MFRReg) remain stable with the increase of labeled target nodes. Moreover, the accuracies of CDNE and AdaGCN slightly increase from $ n=0 $ to $ n=5 $. It indicates these five cross-graph models lack the capability to utilize the target labels well. Note that, under the UDA setting ($ n=0 $), SemiGCL has no an improvement over SemiGCL-DA, since the domain adaptation component of SemiGCL (i.e., minimax entropy training) is devised for the SSDA tasks.
\subsection{Hyperparameter Sensitivity Analysis (RQ4)}
In this section, we analyze the influences of four key hyperparameters on the model performance, including embedding dimension $ l $, contrastive learning coefficient $ \lambda_{1} $, teleport probability $ \alpha $, and temperature parameter $ T $. The aim is to gain insights into how these hyperparameters should be configured. When investigating a specific hyperparameter, the remaining hyperparameters are set to their default values provided in~\ref{appen_imple_details}. Note that, in this section, the labeled set of target graph is identical to the one in Section~\ref{sec_per_sty}.

\begin{figure}[htbp]
	\makebox[\textwidth][c]{\includegraphics[width=12.5cm]{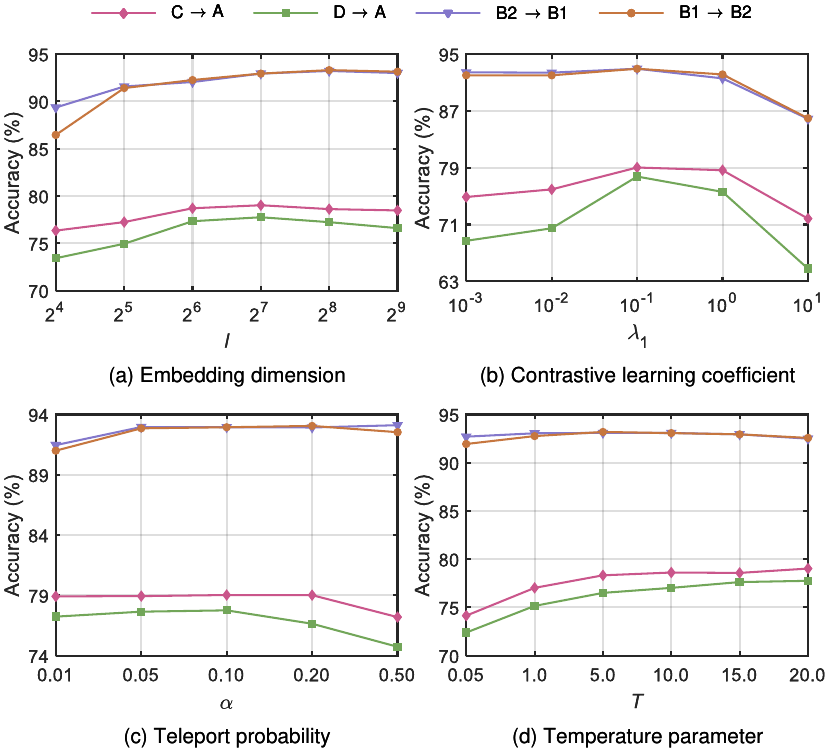}}
	\caption{Classification accuracy varied with each of the four different hyperparameters.}
	\label{fig_hyper_sensi}
\end{figure}

Figure~\ref{fig_hyper_sensi} presents the classification accuracies of transfer tasks on the citation and social graphs. Embedding dimension, $ l $, refers to the dimension of a node representation vector learned by the SemiGCL model. The model achieves the highest accuracy with an embedding dimension of $ 128 $ on the citation graph tasks (i.e., C$ \rightarrow $A and D$ \rightarrow $A). On the social graph tasks (i.e., B2$ \rightarrow $B1 and B1$ \rightarrow $B2), embedding dimensions $ l\in\left\lbrace 128,256,512 \right\rbrace $ can all lead to fairly good performance. Contrastive learning coefficient, $ \lambda_{1} $, is the weight of contrastive loss when optimizing the GNN encoders. As discussed in Section~\ref{sec_model_var}, contrastive learning has a more significant impact on the citation graphs. Therefore, the model performance is more sensitive to the contrastive learning coefficient on transfer tasks C$ \rightarrow $A and D$ \rightarrow $A. When $ \lambda_{1}=0.1 $, the best accuracy is observed on all four transfer tasks. Teleport probability, $ \alpha $, plays an important role in constructing the diffusion matrix. It is observed that the optimal value is around $ \alpha=0.1 $. Temperature parameter, $ T $, scales the output of node classifier. When $ T=5.0 $, the classification accuracy reaches its peak value on transfer tasks B2$ \rightarrow $B1 and B1$ \rightarrow $B2. On the citation graph tasks, a larger temperature parameter (i.e., $ T=20.0 $) is more desirable.
\section{Discussion}
\label{sec_discuss}
SSDA on graphs presents three significant challenges: domain divergence between the source and target graphs, model bias towards the source graph, and non-i.i.d. graph-structured data. Our method, SemiGCL, showcases superior performance on graph SSDA tasks through the principled integration of minimax entropy training and graph contrastive learning. Minimax entropy training plays a crucial role in mitigating domain divergence and alleviating model bias with the entropy loss of unlabeled target nodes. Meanwhile, graph contrastive learning captures both local and global structural information within a graph.

The experiments are conducted on the citation and social graphs, where SemiGCL's performance hinges on the effective functioning of minimax entropy training and graph contrastive learning. On both types of graphs, minimax entropy training consistently improves SemiGCL's performance. Compared with the social graphs, a citation graph is sparser and in a larger scale, increasing the difficulty in capturing global information through local-view neighborhood aggregation alone. Consequently, contrastive learning between the local and global views yields a larger improvement on the citation graphs. On a small-scale and dense social graph, contrastive learning is not that powerful. Additionally, as minimax entropy training is specifically designed for the SSDA tasks, it does not exhibit an improvement under the UDA setting. Hyperparameters are also critical for the performance of SemiGCL, such as embedding dimension, contrastive learning coefficient, teleport probability, and temperature parameter.
\section{Conclusion}
\label{sec_con}
A novel GNN-based model named SemiGCL has been proposed to tackle the semi-supervised domain adaptation problem on graphs. SemiGCL constructs two GNN encoders to extract node representations from two structural views of a graph, i.e., the original graph (local view) and the diffusion-augmented graph (global view). Graph contrastive learning is employed to maximize the mutual information between representations learned from the two structural views. By doing so, the GNN encoders are encouraged to encode rich local and global information in a graph. To mitigate domain discrepancy, the cosine similarity-based node classifier and the GNN encoders are trained in an adversarial manner using the entropy loss of unlabeled target nodes. Experimental results on real-world information networks demonstrate that our method surpasses the state-of-the-art baselines on the benchmark SSDA tasks.

This study considers only a single labeled source graph. Another practical scenario is to have a few labeled source graphs with diverse data distributions. One potential avenue for future research is the development of methods that select the optimal source graph from the available options. Utilizing the optimal source graph, a single-source domain adaption model~\cite{zhao_review_2022} could achieve the best performance on a particular target graph short of labels. Another promising direction for future work is to develop multi-source domain adaptation approaches~\cite{peng_moment_2019} that facilitate the transfer of knowledge from multiple labeled source graphs to the target graph.

\section*{Acknowledgments}
This work was supported in part by the National Natural Science Foundation of China (NSFC) under Grants (62203137, 62362020, 62102124), in part by the Innovation and Technology Commission (ITC) of Hong Kong under Grant MRP/029/20X, in part by the Research Grants Council (RGC) of Hong Kong under Grants (17209021, 17207020, 17210023), and in part by the Alexander von Humboldt Foundation, Germany.
\appendix
\section{Implementation Details}
\label{appen_imple_details}
The proposed SemiGCL\footnote{The source codes of SemiGCL are publicly available at \url{https://github.com/JiarenX/SemiGCL}.} is implemented in PyTorch. In Table~\ref{tab_hyper}, we report the main hyperparameters selected for each transfer task. GNN encoders, $ f_{A} $ and $ f_{P} $, have two-layer structures. The layer dimensions are 1024 and 64 in sequence (i.e., ``1024/64" in Table~\ref{tab_hyper}). Cosine similarity-based classifier, $ f_{c} $, is a logistic regression with $ L_{2} $ normalization applied to its input. The output of node classifier is scaled by a temperature parameter $ T $ and activated by a softmax function. The SemiGCL model is trained over shuffled minibatches using the Adam optimizer. We select initial learning rate $ \eta_{0} $ from $ \left\lbrace 0.005, 0.010, 0.015 \right\rbrace $. When training progress $ p $ linearly increases from $ 0 $ to $ 1 $, we follow~\cite{ganin_domain-adversarial_2016} to decay learning rate as $ \eta_{p} = \eta_{0}\left( 1+10p\right)^{-0.75} $. $ L_{2} $ regularization is imposed on the trainable parameters to prevent overfitting with a weight decay term of $ 5\times10^{-5} $. In the overall objective function (Eq.~\ref{eq_FE_loss}), contrastive learning coefficient, $ \lambda_{1} $, is chosen as $ 0.1 $. Domain adaptation coefficient, $ \lambda_{2} $, starts from $ 0 $ and progressively increases, i.e., $ \lambda_{2} = 2\left( 1+{\rm exp}\left( -10p \right)  \right)^{-1}-1 $. We set the maximum value of $ \lambda_{2} $ as $ 0.1 $. In Eq.~\ref{eq_Cls_loss}, entropy coefficient, $ \lambda_{3} $, is 1.0 for the citation graphs and 2.0 for the social graphs.
\setcounter{table}{0}
\begin{table}[htbp]
	\caption{Main hyperparameters for the SemiGCL model.}
	\label{tab_hyper}
	\centering
 	\scalebox{0.9}{
	\centerline{
	\begin{threeparttable}
		\begin{tabular}{c|c|c|c|c|c|c|c|c}
			\hline
			Transfer Task &     $ \eta_{0} $     &       Epoch        &     Batch Size     &           Weight Decay            &       Layer Dimension       &              $ \bm{s} $\tnote{$ \ast $}              &     $ \alpha $      &        $ T $        \\ \hline
			C$ \rightarrow $A & \multirow{4}*{0.010} & \multirow{4}*{30}  & \multirow{4}*{128} & \multirow{8}*{$ 5\times10^{-5} $} & \multirow{8}*{1024/64} & \multirow{4}*{$\left\lbrace20,20\right\rbrace$} & \multirow{4}*{0.10} & \multirow{6}*{20.0} \\ \cline{1-1}
			D$ \rightarrow $A &                      &                    &                    &                                   &                        &                                                 &                     &                     \\ \cline{1-1}
			A$ \rightarrow $C &                      &                    &                    &                                   &                        &                                                 &                     &                     \\ \cline{1-1}
			D$ \rightarrow $C &                      &                    &                    &                                   &                        &                                                 &                     &                     \\ \cline{1-4}\cline{7-8}
			A$ \rightarrow $D & \multirow{2}*{0.005} & \multirow{2}*{100} & \multirow{2}*{256} &                                   &                        & \multirow{2}*{$\left\lbrace10,10\right\rbrace$} & \multirow{2}*{0.05} &                     \\ \cline{1-1}
			C$ \rightarrow $D &                      &                    &                    &                                   &                        &                                                 &                     &                     \\ \cline{1-4}\cline{7-9}
			B2$ \rightarrow $B1 & \multirow{2}*{0.010} & \multirow{2}*{50}  & \multirow{2}*{64}  &                                   &                        & \multirow{2}*{$\left\lbrace30,30\right\rbrace$} & \multirow{2}*{0.10} & \multirow{2}*{15.0} \\ \cline{1-1}
			B1$ \rightarrow $B2 &                      &                    &                    &                                   &                        &                                                 &                     &                     \\ \hline
		\end{tabular}
		\begin{tablenotes}
			\footnotesize
			\item[$ \ast $] A set $ \bm{s}=\left\lbrace s_{1}, \ldots, s_{K}\right\rbrace  $ consists of the neighborhood sample size $ s_{k} $ at every search depth $ k\in\left\lbrace 1,\ldots, K\right\rbrace$.
		\end{tablenotes}	    	
	\end{threeparttable}}}
\end{table}

The baselines are implemented following the original papers and evaluated with a protocol the same as that of SemiGCL. We adapt baselines in the first category (i.e., GCN, GAT, GraphSAGE, DGI, and MVGRL) to make them conduct cross-graph node classification (Section~\ref{sec_cross_cls_cita} and Section~\ref{sec_cross_cls_social}). The cross-entropy loss is calculated with the labeled nodes in the source and target graphs. For DGI and MVGRL, the contrastive loss is computed independently for each graph, that is, source graph or target graph. The contrastive losses on both graphs are then added to optimize the model. After training, GNN models in the first category are evaluated on the unlabeled nodes of target graph. In single-graph node classification (Section~\ref{sec_sin_cls}), these GNN models are trained and evaluated only with the partially labeled target graph under the semi-supervised learning scenario. DANN, WDGRL, and MME are initially designed to process images with CNNs, such as AlexNet and ResNet34. To encode graph data, their feature extractors are replaced by multilayer perceptrons (MLPs) which take node attributes as input. To adapt DANN and WDGRL for the SSDA tasks, we calculate the cross-entropy loss with the labeled nodes in source and target graphs. 

The cross-graph models are adapted for the SSDA setting. AdaGCN~\cite{dai_graph_2022} is implemented with PyTorch based on the original paper. Although CDNE and AdaGCN also consider the scenario that the target graph is partially labeled, they assume a percentage of target nodes is randomly selected to have labels. We modify the CDNE codes\footnote{\url{https://github.com/shenxiaocam/CDNE}} and the AdaGCN codes to assign the same number of labeled nodes for each class in the target graph. The UDA methods (i.e., ACDNE\footnote{\url{https://github.com/shenxiaocam/ACDNE}}, UDA-GCN\footnote{\url{https://github.com/mandy976/UDAGCN}}, ASN\footnote{\url{https://github.com/yuntaodu/ASN}}, AdaGIn\footnote{\url{https://github.com/JiarenX/AdaGIn}}, MFRReg\footnote{\url{https://github.com/Shen-Lab/GDA-SpecReg}}, and GRADE\footnote{\url{https://github.com/jwu4sml/GRADE}}) are modified for the SSDA setting by incorporating the cross-entropy loss of labeled nodes in the target graph. The hyperparameters of these cross-graph models are initialized with the recommended ones in their papers or official implementations. We have tuned some of their key hyperparameters to improve the performance of these models. For example, the GNN-based models (i.e., AdaGCN, UDA-GCN, ASN, AdaGIn, MFRReg, and GRADE) are tested up to three GNN layers. The dimension of each GNN layer is selected in the set $ \left\lbrace 128, 256, 512, 1024, 2048 \right\rbrace $. 

In all models, except for DGI and MVGRL, embedding dimension $ l $ is set as 128 following ACDNE. Unlike other GNN models that conduct node classification tasks in an end-to-end manner, DGI and MVGRL generate node embeddings first and then train a logistic regression classifier for node classification. We find an embedding dimension of 512 improves the performance of DGI and MVGRL. For each method, we report the mean value of classification accuracies over five runs with different random seeds. Note that, in multiclass classification, according to the scikit-learn document\footnote{\url{https://scikit-learn.org/stable/modules/model_evaluation.html}}, the Micro-Precision, Micro-Recall, and Micro-F1 scores are all identical to the classification accuracy.

\bibliography{./ref/SemiGCL}
\end{document}